\documentclass[letterpaper, 10 pt]{IEEEtran}
\makeatletter
\let\NAT@parse\undefined%
\makeatother
\IEEEoverridecommandlockouts%

\usepackage{amsmath}
\usepackage{amssymb}
\usepackage{balance}
\usepackage{booktabs}    
\usepackage{bm}          
\usepackage[usenames,dvipsnames,table,xcdraw]{xcolor}
\usepackage{pgfplots}
\usepackage{mathrsfs}
\pgfplotsset{compat=1.13}

\usepackage{paralist}
\usepackage{hyperref}
\hypersetup{
	pdffitwindow = true, 	pdftitle     = {root},	pdfauthor    = {Christopher McGreavy},	colorlinks   = true,	
	breaklinks   = true,	urlcolor     = Blue,	linkcolor    = Blue,	citecolor    = Blue
}
\usepackage[all]{hypcap} 
\usepackage[capitalise,nameinlink]{cleveref}
\usepackage{graphicx}
\usepackage[format=default]{subcaption}
\graphicspath{{./figs/}}
\usepackage[ruled,linesnumbered]{algorithm2e}
\usepackage{multirow}
\usepackage{algpseudocode}
\usepackage{siunitx}
\usepackage{adjustbox}
\usepackage{changepage}
\usepackage{xspace}
\usepackage[numbers]{natbib}

\usepackage{tabularx, booktabs} 
\usepackage{multirow}
\newcolumntype{C}[1]{>{\centering\let\newline\\\arraybackslash\hspace{0pt}}m{#1}}
\newcolumntype{L}[1]{>{\raggedright\let\newline\\\arraybackslash\hspace{0pt}}m{#1}}
\newcolumntype{R}[1]{>{\raggedleft\let\newline\\\arraybackslash\hspace{0pt}}m{#1}}
\DeclareMathOperator*{\argmin}{arg\,min}
\DeclareMathOperator*{\argmax}{arg\,max}
\newcommand\subfigexternalwidth{0.32\textwidth}
\newcommand\subfiginternalwidth{\textwidth}
\newcommand\interfigurespace{0.2mm}
\newcommand\trimleft{35mm}
\newcommand\trimbottom{0mm}
\newcommand\trimright{18mm}
\newcommand\trimtop{0mm}
\newcommand\timelinepath[1]{figs/robot_images/snapshots/#1}

\newcommand{\timeline}[3]{%
    \centering
        \begin{subfigure}{0.01\textwidth}%
                \begin{tikzpicture}
                    \definecolor{color1}{rgb}{0.290196078431373,0.525490196078431,0.909803921568627}
                    \node[] at (0,-2) {};
                    \node[] at (5.9,2) {};
                    \draw[gray!30] (0.05,-1.8) rectangle(5.855,1.8);
                    \draw[color1!60, dashed] (2,-1.8) -- (2,1.8);
                    \draw[color1!60, dashed] (3.9,-1.8) -- (3.9,1.8);
                \end{tikzpicture}
        \end{subfigure}
        \begin{subfigure}{\subfigexternalwidth}%
            \includegraphics[width=\subfiginternalwidth,trim = \trimleft{} \trimbottom{} \trimright{} \trimtop{},clip]{\timelinepath{#1}}%
        \end{subfigure}%
        \hspace{\interfigurespace}%
        \begin{subfigure}{\subfigexternalwidth}%
            \includegraphics[width=\subfiginternalwidth,trim = \trimleft{} \trimbottom{} \trimright{} \trimtop{},clip]{\timelinepath{#2}}%
        \end{subfigure}%
        \hspace{\interfigurespace}%
        \begin{subfigure}{\subfigexternalwidth}%
            \includegraphics[width=\subfiginternalwidth,trim = \trimleft{} \trimbottom{} \trimright{} \trimtop{},clip]{\timelinepath{#3}}%
        \end{subfigure}%
    }
    
\title{Reachability Map for Diverse Balancing Strategies and Energy Efficient Stepping of Humanoids}
\author{Christopher McGreavy, Zhibin Li%
\thanks{This research is supported by the EPSRC CDT in Robotics and Autonomous Systems (EP/L016834/1).}
\thanks{The authors are with the School of Informatics, the University of Edinburgh, United Kingdom. Email: {\tt\small c.mcgreavy@ed.ac.uk, zhibin.li@ed.ac.uk}}%
}

\begin{document}
\bstctlcite{IEEEexample:BSTcontrol}
\renewcommand*{\bibfont}{\footnotesize}
\maketitle
\thispagestyle{plain}
\pagestyle{plain}

\begin{abstract}
    In legged locomotion, the relationship between different gait behaviors and energy consumption must consider the full-body dynamics and the robot control as a whole, which cannot be captured by simple models. This work studies the robot dynamics and whole-body optimal control as a coupled system to investigate energy consumption during balance recovery. We developed a 2-phase nonlinear optimization pipeline for dynamic stepping, which generates reachability maps showing complex energy-stepping relations. We optimize gait parameters to search all reachable locations and quantify the energy cost during dynamic transitions, which allows studying the relationship between energy consumption and stepping locations given different initial conditions. We found that to achieve efficient actuation, the stepping location and timing can have simple approximations close to the underlying optimality. Despite the complexity of this nonlinear process, we show that near-minimal effort stepping locations fall within a region of attractions, rather than a narrow solution space suggested by a simple model. This provides new insights into the non-uniqueness of near-optimal solutions in robot motion planning and control, and the diversity of stepping behavior in humans. 
    
\end{abstract}

\section{Introduction}
\label{sec:introduction}
    To achieve long operation time for legged locomotion, it is essential to study the relationship between energy efficiency and locomotion behaviors. The energy efficiency and actuation power of a robot during locomotion are highly related to both its multi-body dynamics and its control, which therefore must be considered together when studying energy consumption.
    
    For high-DoF robots, this hardware and software coupling makes studying energy consumption for general locomotion extremely complex and computationally heavy, given the high dimensionality of the state-action and the solution spaces. In this paper, we study a subset of this problem as a proof of concept to investigate stepping behavior and energy consumption, and shed light on this nonlinear relationship.
    
    \begin{figure}[t]
        \centering
            \def\svgwidth{\columnwidth}
            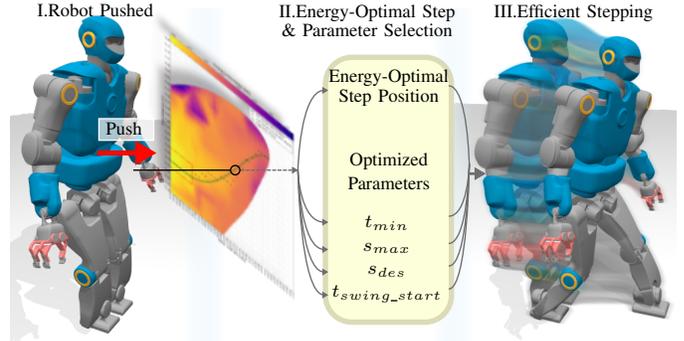
            \caption{
                        Building an energy-optimal reachability map by optimizing parameters of whole-body control for online energy-optimal step selection in balance recovery.
                    }
        \label{fig:posterimage}
    \end{figure}
    
    Even given a reduced complexity of locomotion, such as stepping and balance recovery, the coupling between full-body dynamics and control demands high computation cost due to the large-scale optimization in high-dimensional space. This is the rationale for using simplified models, such as the Linear Inverted Pendulum Model (LIPM) \cite{kajita20013d} for simple step planning \cite{castano2016dynamic}, to reduce the computation for locomotion planning, but they do not capture the complex dynamics of all the robot's links and joints. Therefore, optimal motions based on reduced models are no longer optimal in terms of energy efficiency when considering the whole-body dynamics of a robot.
    
    Due to the disconnect between planning and control, motion plans projected from reduced models onto the whole-body of the robot lose the real optimality of the plan. Moreover, strong nonlinear and coupling effects of dynamics cause deviation between simple model plans and the control during execution, and hence cause errors between predicted and actual energy cost. Therefore, planning and control must be considered together when studying energy consumption, since real optimality depends on the accuracy of control execution given a robot system, making the type of control and its performance an inseparable part of the whole system's energy consumption.
    
    \begin{figure*}[t]
        \centering
            \vspace{-6mm}
            \begin{subfigure}{0.49\textwidth}%
                \centering
                \includegraphics[width=\textwidth]{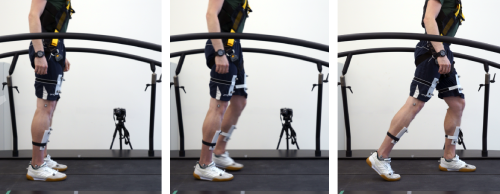}%
                \caption{}
                \label{fig:humansnapshots}
            \end{subfigure}%
            \begin{subfigure}{0.49\textwidth}%
                \centering
                \begin{tikzpicture}

\definecolor{color1}{rgb}{0.290196078431373,0.525490196078431,0.909803921568627}

\begin{axis}[%
width=71.319mm,
at={(0mm,0mm)},
scale only axis,
xmin=0,
xmax=0.5,
xlabel style={font=\color{white!15!black}},
xlabel={Initial CoM Velocity [m/s]},
ymin=-0.01,
ymax=0.6,
ylabel style={font=\color{white!15!black}},
ylabel={Step Position [m]},
axis background/.style={fill=white},
axis x line*=bottom,
axis y line*=left,
height=28.1169375mm,
legend style={legend cell align=left, align=left, draw=none},draw=none,legend style={font=\fontsize{8}{5}\selectfont},ylabel style={font=\footnotesize},xlabel style={font=\footnotesize},tick label style={font=\small},tick label style={font=\footnotesize},yticklabel style={ /pgf/number format/fixed,/pgf/number format/precision=5},scaled y ticks=false
]
\addplot[only marks, mark=*, mark options={}, mark size=1.5000pt, draw=color1, fill=color1, forget plot] table[row sep=crcr]{%
x	y\\
0.104763709288285	0.0080954\\
0.0573119126033036	0.01180076\\
0.053755003194903	0.01347127\\
0.0932314415772038	0.00530588\\
};
\addplot[only marks, mark=*, mark options={}, mark size=1.5000pt, draw=color1, fill=color1, forget plot] table[row sep=crcr]{%
x	y\\
0.0320740803067158	0.00608978000000004\\
0.193866706260814	0.33096458\\
0.138862811873443	0.28637613\\
0.0519187734362963	0.18495094\\
0.0758337468526117	0.19162229\\
0.070307693421579	0.21952182\\
0.0653524170892063	0.18774579\\
0.0381462292209873	0.20319345\\
0.101271123491431	0.2457498\\
0.0496385636419446	0.34153584\\
0.0467348307544268	0.24694796\\
0.063763058483546	0.19076688\\
0.137003513588497	0.21861405\\
0.0802131693057609	0.18463649\\
0.0778369938937743	0.22340127\\
0.0947113283183377	0.1899184\\
0.110658017355788	0.28490717\\
0.270166339400429	0.26136028\\
0.345276819401739	0.28130948\\
0.492809641061774	0.50795993\\
0.397262223498592	0.36078528\\
0.27009789208396	0.26750183\\
};
\end{axis}
\end{tikzpicture}%
%
                \caption{}
                \label{fig:humansteppos}
            \end{subfigure}%
            \caption{
                Human stepping and balance recovery which is hard to model as a biological multi-body system: (a) A subject taking a step during push recovery; (b) Distribution of step locations during various push recovery \cite{mcgreavy2020unified}.
            }
    \end{figure*}
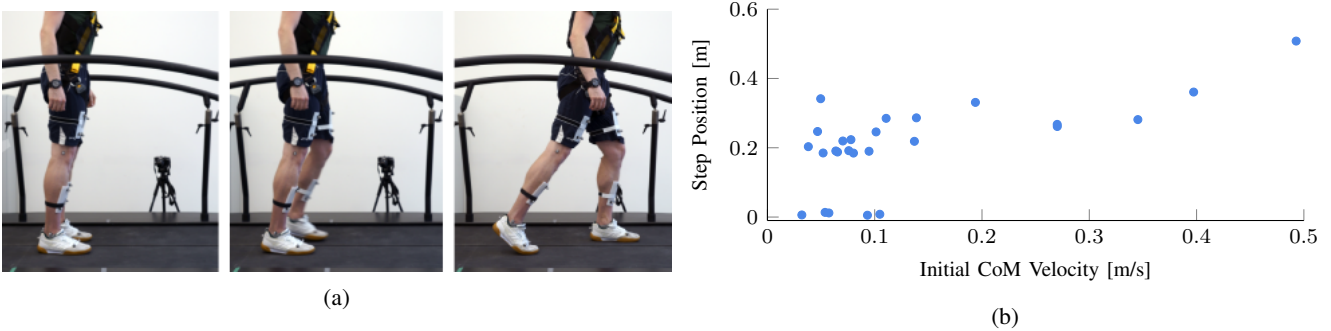
    To plan and control the whole robot, Whole Body Motion Planning (WBM) \cite{dai2014whole} and Whole Body Control (WBC) \cite{khatib2004whole} can be used for constrained planning and control to enforce tasks, such as: kinematic reachability \cite{deits2014footstep,hong2019real}, collision avoidance \cite{baudouin2011real} and realistic centroidal dynamics \cite{aceituno2017simultaneous,ponton2016convex}. Computing WBM plans using Mixed Integer Convex Optimization were studied in \cite{tonneau2020sl1m} \cite{deits2014footstep}, which is intended for online use and has no mechanism for comparing all possibilities and choose energy optimal gaits. Therefore, energy consumption during stepping can be studied using whole-body methods if we expand the search to all feasible gait parameters such as step locations and timings. However, searching the whole solution space is complex and computationally heavy, therefore the study of energy consumption given open gait parameters are investigated by pre-computation and offline search.
    
    In our problem formulation, stepping motions that are pre-computed are used to produce volumetric reachability maps of footstep locations subject to sets of constraints. Previous work produced maps with constraints such as: kinematic reachability \cite{perrin2011fast,fallon2015architecture}, feasible transitions motions \cite{murooka2021humanoid,kuffner2001footstep} and obstacle avoidance \cite{baudouin2011real,kallman2004motion}. Reachability maps can then be used for faster locomotion planning \cite{jorgensen2020finding,burget2015stance}, complex end-pose planning \cite{yang2017efficient}, enabling dynamic transitions \cite{fernbach2018croc} and for learning \cite{lin2020robust}. In this paper, we study reachability maps that encode energy cost for reaching step positions across the whole state space and build simple heuristics to capture a diverse range of efficient stepping motions that can be used to plan motions online with little computational cost.
    
    Though heuristics for energy efficient step regions are nonlinear, they can be learned by humans to achieve complex stepping behaviors. Figure \ref{fig:humansteppos} shows step selection during human push recovery for initial Center of Mass (CoM) push velocities from the work in \cite{mcgreavy2020unified}, which are offset by mean initial velocity of non-stepping trials (\SI{0.1103}{\metre\per\second}) and show the stochasticity of human stepping \cite{kaul2016human,kopitzsch2017optimization}. In this study, our findings on the underlying optimality of stepping motions also help explain the stochasticity of human stepping.
    
    Our paper is motivated to study the relationship between step location and energy optimality for balance recovery from a set of initial conditions, which simple models cannot characterize. We accurately quantify energy cost for reaching the whole stepping space using a full dynamics based physics simulation to fully explore this relationship. We use Bayesian optimization (BO), which is sample efficient for whole-body tasks \cite{yuan2019bayesian}, to optimize open parameters and achieve a wide range of stepping behaviors with different energy cost.
    
    \begin{figure*}[t]
        \centering
        \vspace{-6mm}
        \def\svgwidth{\textwidth}
        \input{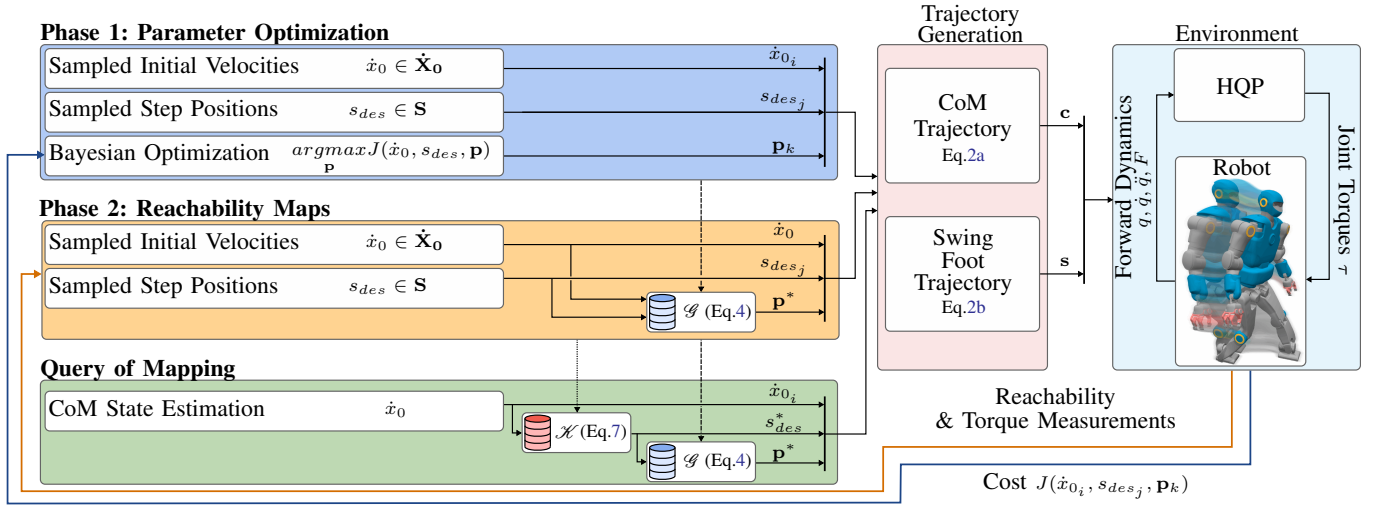}
        \caption{
                A pipeline for optimizing trajectory generation and HQP parameters: Phase 1 optimizes trajectory generation parameters for sets of motions and outputs a parameter mapping; Phase 2 uses this mapping to build reachability maps and an energy-optimal step selection mapping; both mappings are combined for energy-optimal push recovery motions.
            }
        \label{fig:systemoverviewdiagram}
    \end{figure*}
    
    \subsection{Scope}
    \label{sec:scope}
        We develop an optimization pipeline and focus on the following set of stepping motions: taking a single step forward from an initialized standing position with the CoM above stance foot and without toe-off motions. This allows us to search for optimal locomotion parameters given different initial states, and obtain results of the torque usage and energy consumption of a full-body humanoid robot with redundancies.
        
    \subsection{Contributions}
    \label{sec:contributions}
        This paper develops a nonlinear optimization pipeline and studies the efficiency of power consumption in humanoid stepping. Our contributions are as follows:
        \begin{enumerate}
            \item \textbf{Energy-Optimal Stepping (Section \ref{sec:minimumenergystepselection})} Reachability maps that show energy-optimal step positions based on whole-body dynamics and the use of optimal control;
            \item \textbf{Optimization Pipeline (Section \ref{sec:parametertuning}, \ref{sec:feasibilitymaps})} A parallelized optimization pipeline for whole-body control based stepping in full dynamics in a physics simulation; 
            \item \textbf{Reachability Maps (Section \ref{sec:measuredtorquemaps})} A proposed method for computing precise reachability maps for dynamic motions which can rapidly select energy efficient step locations;
            \item \textbf{Finding of Simple Approximation of Optimal Stepping (Section \ref{sec:measuredtorquemaps})} Finding of simple approximations and the disclosure of a funnel of near-optimal step locations.
        \end{enumerate}
        
        By building sampling based energy consumption maps, we reveal the findings of a complex and nonlinear distribution of efficient stepping locations, and show regions of feasibility with similar efficiency, indicating simple heuristics and the non-uniqueness of balance recovery strategies -- which can only be studied and understood by considering the complexity of whole-body dynamics and the optimal control as a whole. We find that such optimal step regions are very different from those predicted based on simple models which do not capture the whole-body dynamics of the system. Moreover, the reachability maps can be used to quickly select step locations and offer insight into explaining human step selections which do not need to repeat the same twice. 
        
        The paper structure is as follows: we mathematically define the pipeline and parameterized control system in Section \ref{sec:pipelineoverview}, then describe the methods of the pipeline in more detail (Section \ref{sec:dynamicsteppingoptimizationpipeline}). We then present our results in section \ref{sec:results}, followed by our discussion (Section \ref{sec:discussion}) and conclusions (Section \ref{sec:conclusion}).

\section{Problem Formulation}
\label{sec:pipelineoverview}
    To study energy efficiency during stepping, we optimize a set of stepping motions to build reachability maps to encode energy consumption during stepping. We achieve this using our 2-phase pipeline, which is defined mathematically below.

    Our pipeline centers on the interaction between parameterized trajectory generators and a Hierarchical Quadratic Programming (HQP) whole-body feedback controller. The HQP, \(\mathscr{H}\), tuned for dynamic stepping, solves joint torques, \(\boldsymbol{\tau}\), given a set of task constraints:
    \begin{equation} \label{eq:hqpdef}
        \mathscr{H} : (\mathbf{c},\mathbf{s},t,\mathbf{q},\mathbf{\dot{q}},\mathbf{\ddot{q}}) \longmapsto \boldsymbol{\tau},
    \end{equation}
    where \((\mathbf{q},\mathbf{\dot{q}},\mathbf{\ddot{q}}) \in \mathbb{R}^m\) are the position, velocity and acceleration of the robot's joints, and \(m\) is the number of joints and \(t\) the current time. The reference trajectories \(\mathbf{c}\) and \(\mathbf{s}\) are for the CoM and the swing foot respectively. Full details of the HQP are given in Section \ref{sec:hqpstructure}.
    
    Trajectories are time indexed position references such that \(\mathbf{c}(t) \in \mathbb{R}^3\) and \(\mathbf{s}(t) \in \mathbb{R}^3\), generated by the parameterized functions:
    \begin{subequations}
        \begin{align}
            \mathscr{C}_{traj\_gen} :\: (t_{min},s_{max}) \longmapsto \mathbf{c} \label{eq:comtrajdef}\\
            \mathscr{S}_{traj\_gen} :\: (s_{des},t_{swing\_start},s_{speed}) \longmapsto \mathbf{s}. \label{eq:swingtrajdef}
        \end{align}
    \end{subequations}
    CoM trajectory generator \(\mathscr{C}_{traj\_gen}\) takes minimum step time, \(t_{min}\), and maximum step length, \(s_{max}\), as arguments. Swing foot trajectory generator \(\mathscr{S}_{traj\_gen}\) is a function of desired step position \(s_{des}\), swing start time \(t_{swing\_start}\) and step speed \(s_{speed}\) parameters. Parameters are summarized in Table \ref{tab:optvariables}. Phase 1 of our pipeline optimizes these open parameters to produce stepping motions. Open parameters are concatenated in vector \(\mathbf{p}\):
    \begin{equation} \label{eq:paramvectordef}
        \mathbf{p} = [t_{min},s_{max},t_{swing\_start},s_{speed}]^T.
    \end{equation} 
    The output of phase 1 is the mapping \(\mathscr{G}\) from initial CoM velocity, \(\dot{x}_0\), and desired step position, \(s_{des}\), to a set of optimal parameters \(\mathbf{p}^*\) (optimal values are denoted by \((\cdot)^*\)), defined as:
    \begin{equation} \label{eq:mappingfunctiondef}
        \mathscr{G} : (\dot{x}_0,s_{des}) \longmapsto \mathbf{p^*} = \underset{\mathbf{p}}{argmax} \quad J(\dot{x}_0,s_{des},\mathbf{p}),
    \end{equation}
    such that maximizing this objective function will minimize all the error terms.
    To build this mapping, we maximize the objective function \(J\) for pairs of initial conditions \(\dot{x}_0\) and \(s_{des}\) using BO.
    
    \begin{table}[t]
        \centering
        \caption{Objective function weights and their values.}
        \label{tab:objectiveweightstable}
            \begin{tabular}{@{}lll@{}}
        \toprule
        Notation        & Affects            & Value \\ \midrule
        \(w_{f}\)       & Failure            & 0.001  \\
        \(w_{swing}\)   & Swing Leg Position error    & 50     \\
        \(w_{x\_mid}\)  & Final CoM Position & 1     \\
        \(w_{z}\)       & Final CoM Height   & 1     \\
        \(w_{\tau}\)    & Torque Consumption & 0.0002  \\ \bottomrule
    \end{tabular}
        \vspace{-3mm}
    \end{table}
    
    At each BO iteration, the initial conditions and BO parameters are passed to the trajectory generators (Eq.\ref{eq:comtrajdef}, \ref{eq:swingtrajdef}), which are then passed as input to the HQP (Eq.\ref{eq:hqpdef}), which interfaces with a full dynamic simulation environment where the robot attempts to step towards the desired step position \(s_{des}\). This is highlighted in blue in Figure \ref{fig:systemoverviewdiagram}). The objective function \(J\) is a function of values from the dynamic simulation, which we consider to be a black-box and each term is assigned a manually tuned weight, each denoted by a subscript of \(w\), defined in Table \ref{tab:objectiveweightstable}:
    \begin{equation}\label{eq:costfunction}
        \begin{aligned}
            J(\dot{x}_0,s_{des},\mathbf{p}) &= -(w_f (t_{total}-t_{term}) \\
            + &w_{swing} (s_{des} - s_{td})^2 + w_{x\_mid} (x_{f} - s_{mid})^2 \\
            + &w_{z} (z_{nom} - z_{f})  + w_{\tau} J_\tau(\dot{x}_0,s_{des},\mathbf{p})),
        \end{aligned}
    \end{equation}
    The first term applies a cost for early termination, where termination time \(t_{term}\) is less than total simulation time \(t_{total}\), indicating that stepping failed. The value of \(t_{total}\) is predefined and is set to \SI{7}{\second} in this paper. Simulations are terminated either when time \(t = t_{total}\) and the robot is stable having completed the motion, or at \(t_{term}\) if the robot falls (more details in Section \ref{sec:experimentalsetup}).
    
    The difference between the swing foot position at touchdown, \(s_{td}\), and the desired step location, \(s_{des}\), incurs a cost proportional to the error between them. To increase stability after the step, a cost is applied between final CoM position, \(x_{f}\), and the midpoint between the swing and the stance foot: \(s_{mid} = s_{stance}+(s_{stance} - s_{td})\). To encourage straight legs after landing, a cost is applied to final CoM height \(z_{f}\) to be as close as possible to a nominal height \(z_{nom}=\SI{0.925}{\metre}\). A term is added to minimize the integral of measured torque \(\boldsymbol{\tau}\) in all the robot's joints between the time swing foot motion begins \(t_{lo}\) and touchdown time \(t_{td}\):
    \begin{equation}\label{eq:torqueobjectivedef}
        J_\tau : (\dot{x}_0,s_{des},\mathbf{p}) \longmapsto \int_{t_{lo}}^{t_{td}} \boldsymbol{\tau}^2 \; dt.
    \end{equation}
    
    After BO is completed for each pair of initial conditions, we are left with the mapping function \(\mathscr{G}\) (Eq.\ref{eq:mappingfunctiondef}), which outputs optimized parameters for pairs of initial conditions. In phase 2, we query this mapping with a larger set of initial velocities and desired step positions to test how the optimized parameters generalize to novel initial condition pairs. Here, the objective function \(J\) is no longer used, as parameters have already been optimized and stored in the mapping function; instead, the dynamic simulation returns two values, shown in Figure \ref{fig:systemoverviewdiagram}: a binary reachability value to encode whether motions are successful, and the integral of the measured torque, \(J_\tau\) (Eq.\ref{eq:torqueobjectivedef}), during the swing phase for every pair of initial conditions.
    \begin{figure}[t]
                \centering
                \input{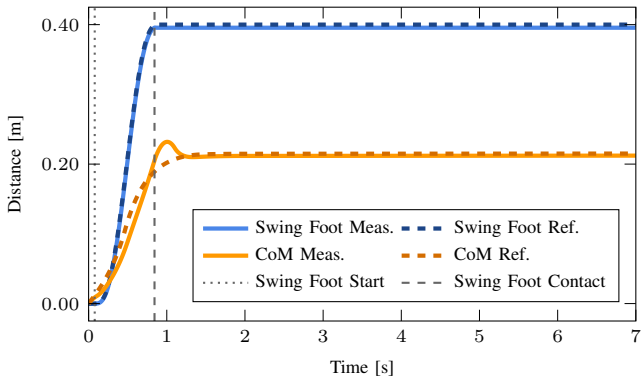}
                \caption{
                        Example reference trajectories for the CoM and swing foot, and measured trajectories from the HQP. Initial conditions: \(\dot{x}_{0}\!=\)\SI{0.15}{\metre\per\second}, \(s_{des}\!=\SI{0.4}{\metre}\).
                        }
                \label{fig:trajectoryplots}
    \end{figure} 
    
    From these values we build a map to encode which step positions are reachable from each initial CoM velocity and a map to quantify the measured torque during the swing phase of each reachable step position. Using this measured torque map, we create a second mapping, \(\mathscr{K}\), from the robot's initial CoM velocity \(\dot{x}_0\) to an energy-optimal step position \(s_{des}^*\).
    
    Since \(\mathscr{K}\) automatically generates energy-optimal step positions \(s_{des}^*\), it is combined with mapping function \(\mathscr{G}\) to reduce the input dimensionality such that only initial CoM velocity \(\dot{x}_0\) is required to generate optimal trajectory generation parameters \(\mathbf{p}^*\) to reach the energy-optimal step position \(s_{des}^*\): 
    
    \begin{flalign}\label{eq:distilledmappingfunction}
        \setlength{\thinmuskip}{0mu} \setlength{\medmuskip}{0mu} \setlength{\thickmuskip}{0mu}
        \mathscr{K}:(\dot{x}_0) \mapsto
            \begin{cases}
                \mathbf{p}^* = \mathscr{G}(\dot{x}_0,s_{des}^*)\\
                \text{\small{s.t.:}} s_{des}^* = \underset{\quad\quad s_{des}}{\argmin} \; J_\tau(\dot{x}_0,s_{des},\mathscr{G}(\dot{x}_0,s_{des})).
            \end{cases}
    \end{flalign}

    The maps are built offline and a regression model is used to approximate energy-optimal step positions \(s_{des}^*\) for arbitrary initial CoM velocities \(\dot{x}_0\). After this pipeline is complete, we can query these mappings to execute push recovery motions, and we run dynamic simulations validation tests using the mappings and previously unseen initial conditions. We next define the control structure of the low-level HQP controller and the trajectory generation modules.

    \subsection{HQP Structure}
    \label{sec:hqpstructure}
        The HQP feedback controller is a function of forward dynamics and two parameterized trajectories (Eq.\ref{eq:hqpdef}). We tuned the tasks, weights and hierarchy order of an existing HQP Controller \cite{tsidprete}. It is worth noting that dynamic stepping is restricted if the CoM task has higher priority than the swing foot task, instead, CoM tasks should be at the same priority, or lower, than the swing foot.
        
    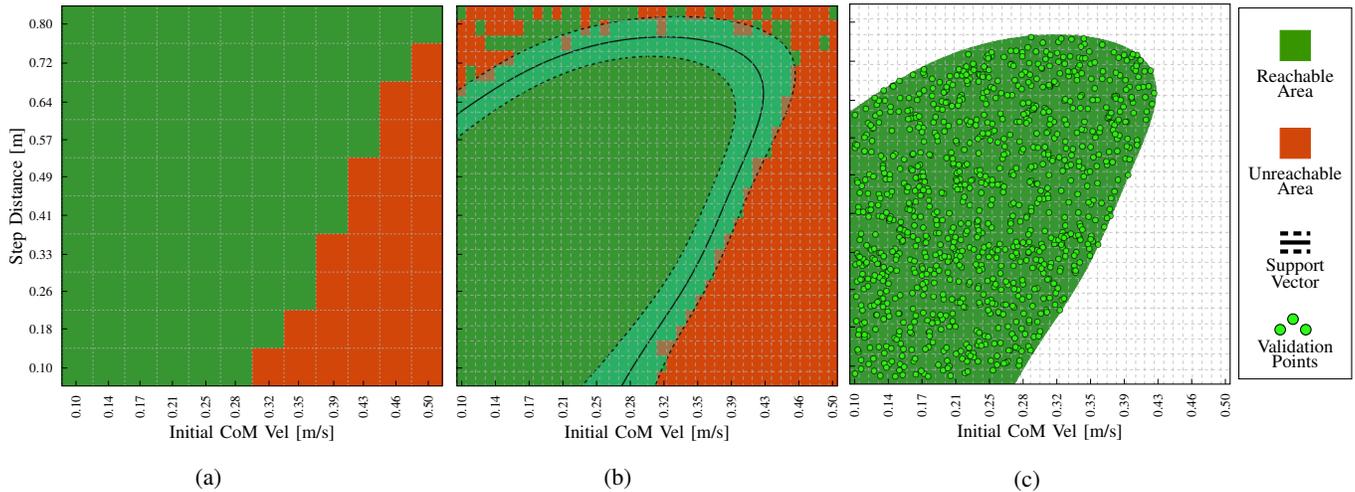
\begin{figure*}[t]
        \centering
            \vspace{-6mm}
            \begin{subfigure}{0.3\textwidth}%
\raisebox{0mm}{
\hspace{-2.9mm}
\begin{tikzpicture}
\begin{axis}[
axis on top,
font=\tiny,
height=1.22\textwidth,
width=1.22\textwidth,
xmin=0.09487179485000001, xmax=0.50512820515,
xminorgrids,
xtick pos=left,
xtick style={color=black},
xtick={
0.11,
0.145,
0.179,
0.214,
0.25,
0.28181818181818186,
0.31818181818181823,
0.35,
0.388,
0.42,
0.4549999,
0.49
},
xticklabel style = {rotate=90.0},
xticklabels={
0.10,
0.14,
0.17,
0.21,
0.25,
0.28,
0.32,
0.35,
0.39,
0.43,
0.46,
0.50
},
ylabel={Step Distance [m]},
ymin=0.0854166666666667, ymax=0.814583333333333,
yminorgrids,
ytick pos=left,
ytick style={color=black},
every major tick/.append style={major tick length=2pt, black},
ytick={
0.12,
0.195,
0.267,
0.338,
0.413,
0.487,
0.558,
0.63,
0.705,
0.78  
},
xlabel={Initial CoM Vel [m/s]},
x label style={at={(0.5,-0.08)}},
xlabel style={font=\scriptsize},
ylabel style={font=\scriptsize},
yticklabel style={/pgf/number format/fixed,/pgf/number format/precision=2,/pgf/number format/fixed zerofill}, scaled y ticks = false,,
y label style={at={(-0.07,0.5)}},
yticklabels={
0.10,
0.18,
0.26,
0.33,
0.41,
0.49,
0.57,
0.64,
0.72,
0.80
}
]
\addplot graphics [includegraphics cmd=\pgfimage,xmin=0.09487179485000001, xmax=0.50512820515, ymin=0.0854166666666667, ymax=0.814583333333333] {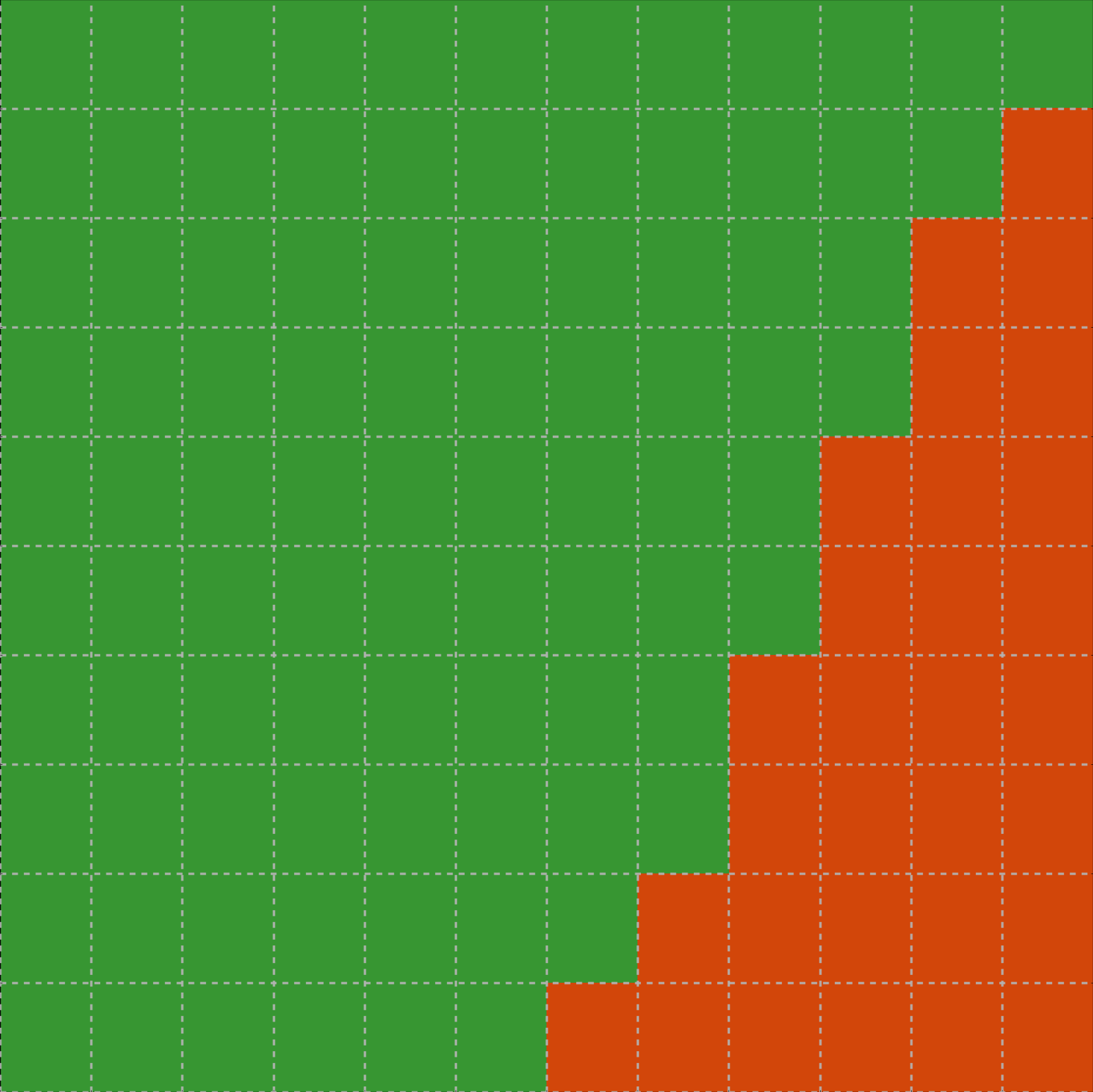};
\end{axis}
\end{tikzpicture}
}

       


                \caption{}
                \label{fig:parammap}
            \end{subfigure}%
            \begin{subfigure}{0.3\textwidth}%
                \raisebox{0.0mm}{
\hspace{0.95mm}
\begin{tikzpicture}
\begin{axis}[
axis on top,
font=\tiny,
height=1.22\textwidth,
width=1.22\textwidth,
xmin=0.09487179485000001, xmax=0.50512820515,
xminorgrids,
xtick pos=left,
xtick style={color=black},
xtick={
0.1,
0.13636364,
0.17272727,
0.20909091,
0.24545455,
0.28181818,
0.31818182,
0.35454545,
0.39090909,
0.42727273,
0.46363636,
0.5 
},
xticklabel style = {rotate=90.0},
xticklabels={
0.10,
0.14,
0.17,
0.21,
0.25,
0.28,
0.32,
0.35,
0.39,
0.43,
0.46,
0.50
},
ymin=0.0854166666666667, ymax=0.814583333333333,
yminorgrids,
xlabel={Initial CoM Vel [m/s]},
x label style={at={(0.5,-0.08)}},
xlabel style={font=\scriptsize},
ytick pos=left,
ytick style={color=black},
every major tick/.append style={major tick length=2pt, black},
ytick={
0.12,
0.195,
0.267,
0.338,
0.413,
0.487,
0.558,
0.63,
0.705,
0.78  
},
yticklabel style={/pgf/number format/fixed,/pgf/number format/precision=2,/pgf/number format/fixed zerofill}, scaled y ticks = false,,
yticklabels={ , , , , , , , }
]
\addplot graphics [includegraphics cmd=\pgfimage,xmin=0.09487179485000001, xmax=0.50512820515, ymin=0.0854166666666667, ymax=0.814583333333333] {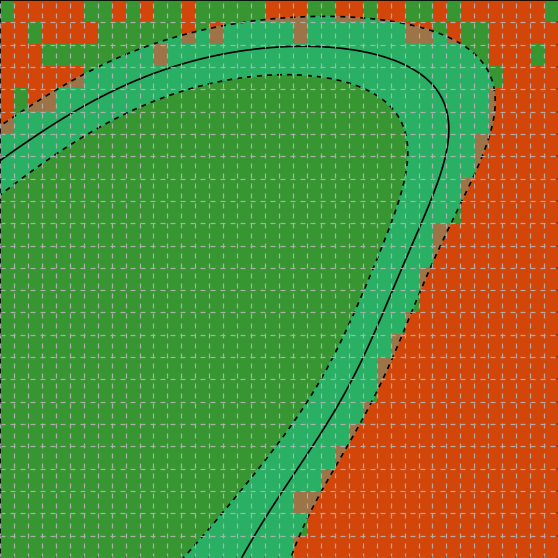};
\end{axis}

\end{tikzpicture}
}%
                \caption{}
                \label{fig:svmmap}
            \end{subfigure}%
            \begin{subfigure}{0.3\textwidth}%
\raisebox{0.5mm}{
\begin{tikzpicture}
\begin{axis}[
axis on top,
font=\tiny,
height=1.22\textwidth,
width=1.22\textwidth,
xmin=0.09487179485000001, xmax=0.50512820515,
xminorgrids,
xtick pos=left,
xtick style={color=black},
xtick={
0.1,
0.13636364,
0.17272727,
0.20909091,
0.24545455,
0.28181818,
0.31818182,
0.35454545,
0.39090909,
0.42727273,
0.46363636,
0.5 
},
xticklabel style = {rotate=90.0},
xticklabels={
0.10,
0.14,
0.17,
0.21,
0.25,
0.28,
0.32,
0.35,
0.39,
0.43,
0.46,
0.50
},
ymin=0.0854166666666667, ymax=0.814583333333333,
yminorgrids,
ytick pos=left,
xlabel={Initial CoM Vel [m/s]},
x label style={at={(0.5,-0.08)}},
xlabel style={font=\scriptsize},
ytick style={color=black},
every major tick/.append style={major tick length=2pt, black},
ytick={
0.12,
0.195,
0.267,
0.338,
0.413,
0.487,
0.558,
0.63,
0.705,
0.78  
},
yticklabel style={/pgf/number format/fixed,/pgf/number format/precision=2,/pgf/number format/fixed zerofill}, scaled y ticks = false,,
yticklabels={ , , , , , , , }
]
\addplot graphics [includegraphics cmd=\pgfimage,xmin=0.09487179485000001, xmax=0.50512820515, ymin=0.0854166666666667, ymax=0.814583333333333] {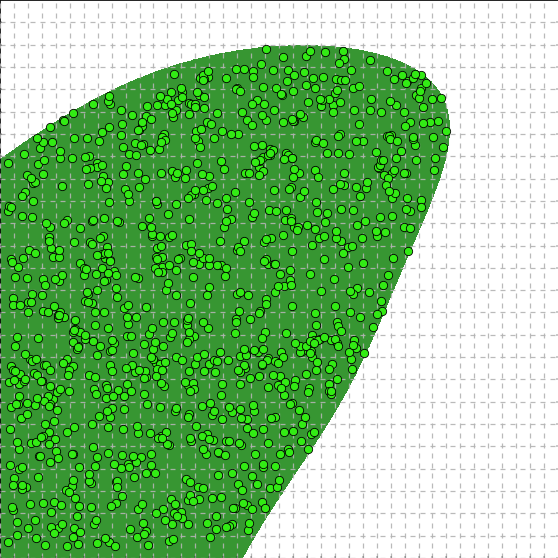};
\end{axis}

\end{tikzpicture}
}%
                \caption{}
                \label{fig:reachvalidationmap}
            \end{subfigure}%
            \begin{subfigure}{0.1\textwidth}
            \centering
                \begin{tikzpicture}
    \definecolor{reachcol}{rgb}{0.21568627450980393,0.5882352941176471,0}
    \definecolor{nonreachcol}{rgb}{0.8235294117647058,0.27450980392156865,0.0392156862745098}
    \definecolor{validationcol}{rgb}{0.1568627450980392, 1.0, 0.0784313725490196}
    
    \node[] at (0,-0.5) {};
    \draw[] (-0.55,0.865) rectangle(0.95,5.8);
    \filldraw[reachcol](0,5.1) rectangle (0.4,5.5);
    \node[] at (0.2,4.9) {\scriptsize{Reachable}};
    \node[] at (0.2,4.7) {\scriptsize{Area}};
    
    \filldraw[nonreachcol](0,3.8) rectangle (0.4,4.2);
    \node[] at (0.2,3.6) {\scriptsize{Unreachable}};
    \node[] at (0.2,3.4) {\scriptsize{Area}};
    
    \draw[ultra thick, black, densely dashed](0,2.8) -- (0.4,2.8);
    \draw[ultra thick, black](0,2.68) -- (0.4,2.68);
    \draw[ultra thick, black, densely dashed](0,2.56) -- (0.4,2.56);
    
    \node[] at (0.2,2.36) {\scriptsize{Support}};
    \node[] at (0.2,2.16) {\scriptsize{Vector}};
    
    \filldraw[color=black,fill=validationcol](0,1.52) circle (2pt);
    \filldraw[color=black,fill=validationcol](0.17,1.66) circle (2pt);
    \filldraw[color=black,fill=validationcol](0.34,1.52) circle (2pt);
    \node[] at (0.2,1.3) {\scriptsize{Validation}};
    \node[] at (0.2,1.1) {\scriptsize{Points}};
    
\end{tikzpicture}
            \end{subfigure}
            \caption{
                        Binary reachability maps show successful (green) or unsuccessful (red) steps for pairs of initial conditions: (a) Parameter optimization map, (b) Dense reachability map, (c) SVM generated reachability map with validation testing points.
                    }
    \end{figure*}
    \subsection{Trajectory Generation}
    \label{sec:trajectorygeneration}
        CoM and swing foot trajectory generation determine the motion produced by the HQP. Each has 2 open parameters which determine their profile, and outputs time indexed position references for the \(X\),\(Y\) and \(Z\) axes. Parameters determine the length of the trajectory and its gradient. Phase 1 of our pipeline (Section \ref{sec:parametertuning}), optimizes parameters for both trajectory generators for every pair of sampled initial conditions, which creates the mapping \(\mathscr{G}\).
        \begin{table}[t]
            \centering
            \caption{Optimization variables and descriptions.}
            \label{tab:optvariables}
            \begin{tabular}{lccccc}
\hline
\begin{tabular}[c]{@{}l@{}}Optimization\\ Parameter\end{tabular} & Description                                                      & Affects                                              & Bounds                                               & \begin{tabular}[c]{@{}c@{}}Optimized\\ Range\end{tabular} & Unit                       \\ \hline
\(t_{min}\)                                                      & \begin{tabular}[c]{@{}c@{}}Minimum\\swing time\end{tabular}     & CoM                                                  & \begin{tabular}[c]{@{}c@{}}0.01-\\ 0.99\end{tabular} & \begin{tabular}[c]{@{}c@{}}0.01-\\ 0.77\end{tabular}      & \SI{}{\second}           \\
\(s_{max}\)                                                      & \begin{tabular}[c]{@{}c@{}}Maximum\\step length\end{tabular}    & CoM                                                  & \begin{tabular}[c]{@{}c@{}}0.01-\\ 0.99\end{tabular} & \begin{tabular}[c]{@{}c@{}}0.06-\\ 0.98\end{tabular}      & \SI{}{\meter}            \\
\(t_{swing\_start}\)                                             & \begin{tabular}[c]{@{}c@{}}Swing foot \\start time\end{tabular} & \begin{tabular}[c]{@{}c@{}}Swing\\ Foot\end{tabular} & \begin{tabular}[c]{@{}c@{}}0.01-\\ 0.08\end{tabular} & \begin{tabular}[c]{@{}c@{}}0.026-\\ 0.078\end{tabular}    & \SI{}{\second}           \\
\(s_{swing\_speed}\)                                             & \begin{tabular}[c]{@{}c@{}}Swing foot\\velocity\end{tabular} & \begin{tabular}[c]{@{}c@{}}Swing\\ Foot\end{tabular} & \begin{tabular}[c]{@{}c@{}}0.2-\\ 3.0\end{tabular}   & \begin{tabular}[c]{@{}c@{}}0.2-\\ 1.34\end{tabular}       & \SI{}{\meter\per\second} \\ \hline
\end{tabular}
        \end{table}
        \subsubsection{CoM Trajectory Generation}
        \label{sec:comtrajectorygeneration}
            An existing LIPM based model \cite{hu2018comparison} was used to produce CoM trajectories and is a function of two parameters: minimum step time \(t_{min}\) and maximum step length \(s_{max}\). An example trajectory is shown in Figure \ref{fig:trajectoryplots}. Originally, the parameter \(s_{max}\) was intended to reflect the capabilities of the real robot, but this results in falling, so this parameter was added to the optimization. Since this pipeline is modular, this can be replaced with alternative CoM trajectory generation methods.

        \subsubsection{Swing Foot Trajectory Generation}
        \label{sec:swingfoottrajectorygeneration}
            Swing foot trajectories are \(5^{\text{th}}\) degree minimum-jerk polynomials \cite{flash1985coordination} (Figure \ref{fig:trajectoryplots}), parameterized by the time at which the swing foot starts to move, (\(t_{swing\_start}\)), and the swing foot speed (\(s_{speed}\)). The Z axis consists of two minimum jerk trajectories connected at a via point at the maximum desired swing height \(z_{max}\), for which we found \SI{80}{\milli\metre} to be a reliable value.

\section{Building Reachability Map: Dynamic Stepping Optimization Pipeline}
\label{sec:dynamicsteppingoptimizationpipeline}
    \subsection{Simulation Setup}
    \label{sec:simulationsetup}
        For the dynamic simulation, we used Pinocchio dynamics library \cite{carpentier2019pinocchio}. In the long term we aim to move the query of mapping implementation onto the real robot, but was not possible for this paper. This pipeline is compatible with any humanoid robot, here we used the Talos humanoid robot model (32 DoF), with the complete dynamic and kinematic properties of the real robot, including the position, velocity, acceleration and torque limits. The simulation environment was fully dynamic, including friction, torque limits, with a simulation frequency of \SI{1000}{\hertz} (\(\Delta t=\SI{0.001}{\second}\)). The \texttt{BayesianOptimization} package \cite{bayesoptpython} was used for parameter optimization.
    
    \subsection{Experimental Setup}
    \label{sec:experimentalsetup}
        During this proof of concept stage, we constrain motions to gait initiation in the \(X\) axis (forward), where in each simulation episode the robot starts in a standing configuration with the swing foot \SI{1}{\centi\metre} above the ground. We start with the swing foot off the ground to avoid the added complexity of optimizing weight transfer time while we develop the pipeline. 
        
        Initial CoM velocities are achieved by inducing reference torques directly at the joints in simulation, torque values are calculated using the Jacobian from the stance foot to the CoM. On the real robot, we expect CoM velocities to be applied by having the robot lean in a given direction or being pushed. Termination conditions during optimization are as follows: robot reaches desired foot position and remains standing at \(t=t_{total}\) (success) or the norm of joint velocities exceed a threshold (\(1e6\)) (failure). If \(t_{f}<t_{total}\), the remaining sensor readings are filled with a nominal high value.
    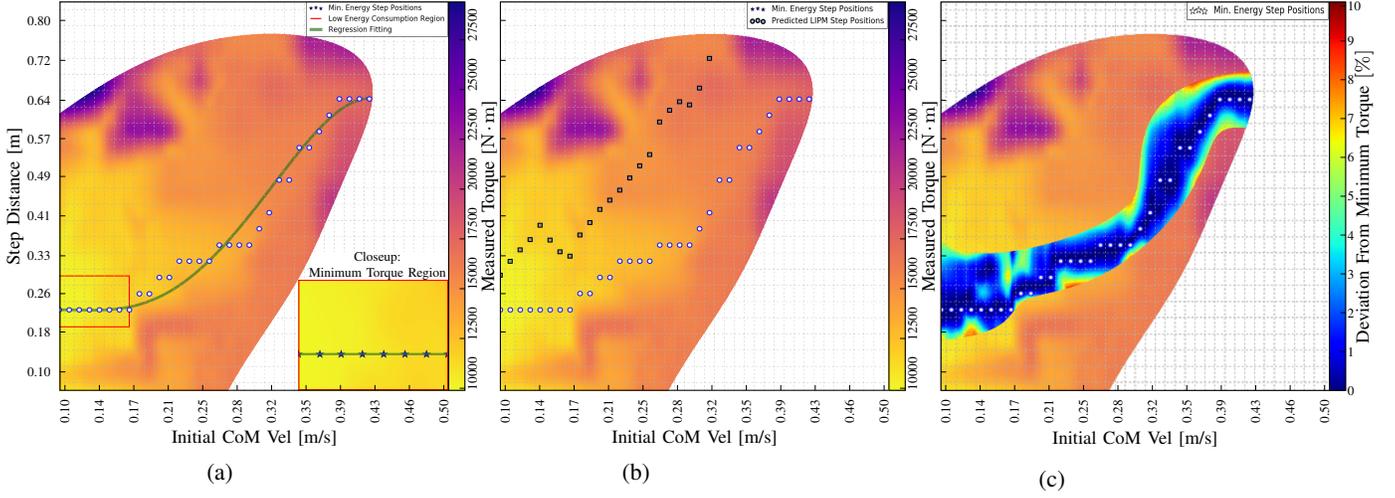
\begin{figure*}[t]
            \vspace{-6mm}
            \begin{subfigure}{0.305\textwidth}%
                \centering
\raisebox{0mm}{
\hspace{-4.3mm}
\begin{tikzpicture}
\definecolor{color0}{rgb}{0.219607843137255,0.462745098039216,0.113725490196078}
\begin{axis}[
name=plot1,
axis on top,
font=\tiny,
height=1.22\textwidth,
width=1.22\textwidth,
xmin=0.09487179485000001, xmax=0.50512820515,
xminorgrids,
xtick pos=left,
xtick style={color=black},
xtick={
0.1,
0.13636364,
0.17272727,
0.20909091,
0.24545455,
0.28181818,
0.31818182,
0.35454545,
0.39090909,
0.42727273,
0.46363636,
0.5 
},
xticklabel style = {rotate=90.0},
xticklabels={
0.10,
0.14,
0.17,
0.21,
0.25,
0.28,
0.32,
0.35,
0.39,
0.43,
0.46,
0.50
},
xlabel={Initial CoM Vel [m/s]},
x label style={at={(0.5,-0.08)}},
xlabel style={font=\scriptsize},
ylabel style={font=\scriptsize},
ylabel={Step Distance [m]},
y label style={at={(-0.07,0.5)}},
ymin=0.0854166666666667, ymax=0.814583333333333,
yminorgrids,
ytick pos=left,
ytick style={color=black},
every major tick/.append style={major tick length=2pt, black},
ytick={
0.12,
0.195,
0.267,
0.338,
0.413,
0.487,
0.558,
0.63,
0.705,
0.78  
},
yticklabel style={/pgf/number format/fixed,/pgf/number format/precision=2,/pgf/number format/fixed zerofill}, scaled y ticks = false,,
yticklabels={
0.10,
0.18,
0.26,
0.33,
0.41,
0.49,
0.57,
0.64,
0.72,
0.80
}
]
\addplot graphics [includegraphics cmd=\pgfimage,xmin=0.09487179485000001, xmax=0.50512820515, ymin=0.0854166666666667, ymax=0.814583333333333] {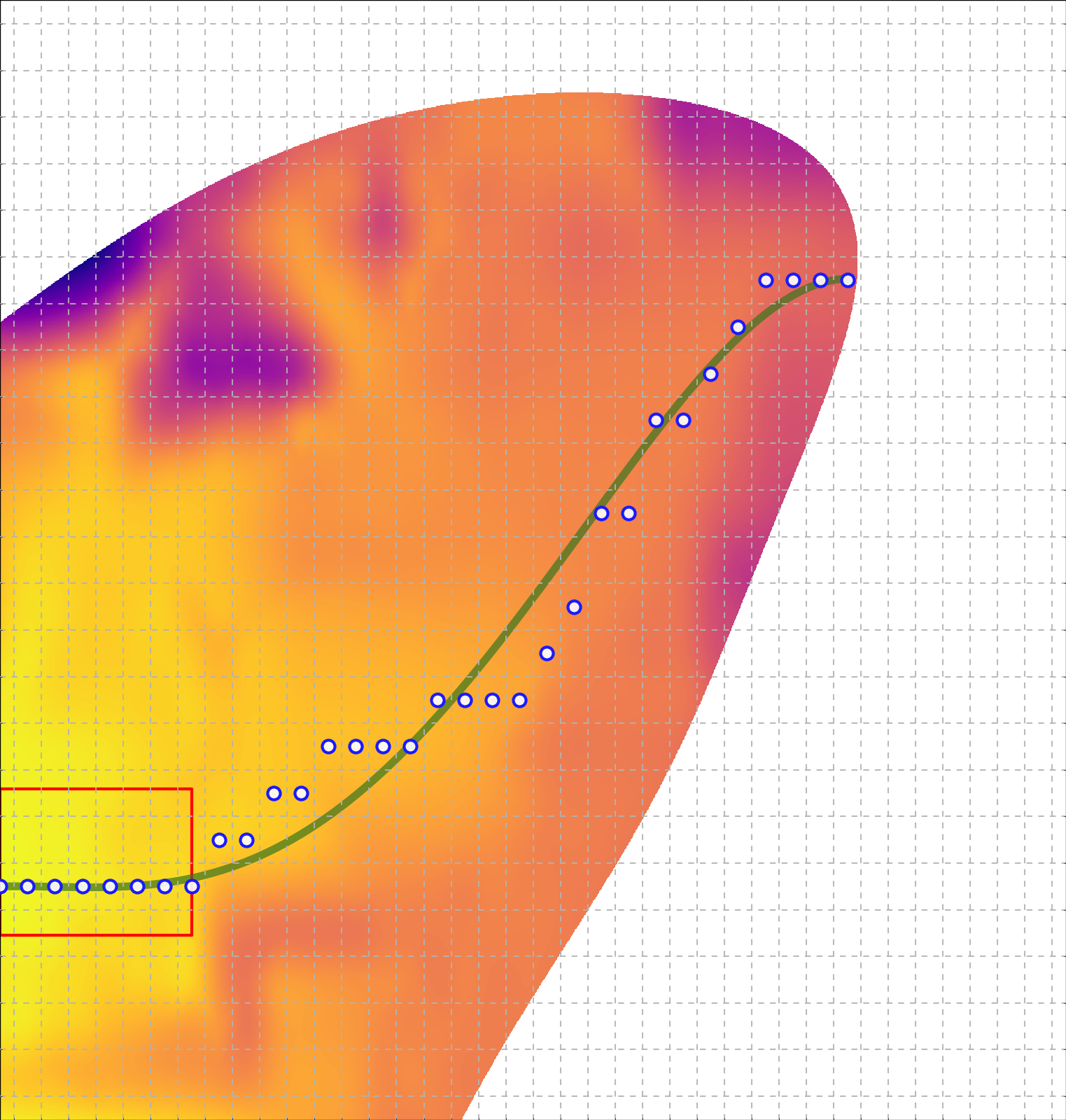};

\filldraw[fill=white,draw=white] (axis cs: 0.362,0.292) rectangle (axis cs: 0.5041,0.344);
\node[align=center] at (axis cs: 0.430,0.319) {Closeup:\\Minimum Torque Region};
\addplot graphics [includegraphics cmd=\pgfimage,xmin=0.347, xmax=0.5036, ymin=0.0854166666666667, ymax=0.292] {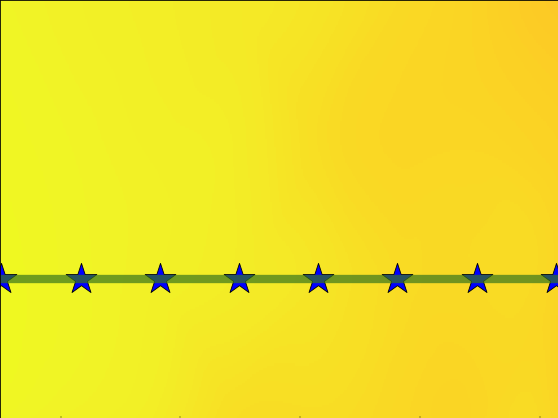};
\addplot graphics [includegraphics cmd=\pgfimage,xmin=0.355, xmax=0.5036, ymin=0.75, ymax=0.81458333] {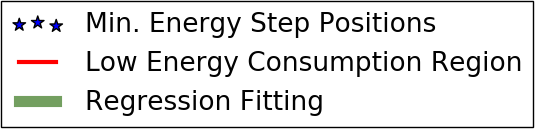};
\draw[draw=red] (axis cs: 0.34702,0.0865) rectangle (axis cs: 0.5041,0.292);
\end{axis}

\begin{axis}[
at=(plot1.south east),
anchor=south west,
axis on top,
font=\tiny,
width=0.325\textwidth,
height=1.22\textwidth,
xmin=0, xmax=1,
ymin=0, ymax=1,
yminorgrids,
ytick pos=right,
xmajorticks=false,
ytick style={color=black},
ytick={
0.00530573,
0.13351371,
0.26172169,
0.38992968,
0.51813766,
0.64634564,
0.77455362,
0.9027616
},
ylabel style={font=\scriptsize},
ylabel={Measured Torque [\SI{}{\newton\metre}]},
y label style={at={(1.45,0.5)}},
yticklabel style = {rotate=90.0,yshift=0.85mm,xshift=2mm}, 
yticklabels={
10000,
12500,
15000,
17500,
20000,
22500,
25000,
27500 
},
every major tick/.append style={major tick length=2pt, black},
]
\addplot graphics [includegraphics cmd=\pgfimage,xmin=0, xmax=1, ymin=0, ymax=1] {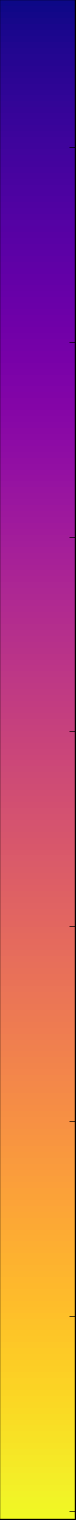};
\end{axis}
\end{tikzpicture}
}
                \vspace{-5.5mm}
                \caption{}
                \label{fig:torquemap}
            \end{subfigure}%
            \begin{subfigure}{0.305\textwidth}%
                \centering
\raisebox{0.0mm}{
\hspace{4.75mm}
\begin{tikzpicture}
\begin{axis}[
name=plot1,
axis on top,
font=\tiny,
height=1.22\textwidth,
width=1.22\textwidth,
xmin=0.09487179485000001, xmax=0.50512820515,
xminorgrids,
xtick pos=left,
xtick style={color=black},
xtick={
0.1,
0.13636364,
0.17272727,
0.20909091,
0.24545455,
0.28181818,
0.31818182,
0.35454545,
0.39090909,
0.42727273,
0.46363636,
0.5 
},
xticklabel style = {rotate=90.0},
xticklabels={
0.10,
0.14,
0.17,
0.21,
0.25,
0.28,
0.32,
0.35,
0.39,
0.43,
0.46,
0.50
},
xlabel={Initial CoM Vel [m/s]},
x label style={at={(0.5,-0.08)}},
xlabel style={font=\scriptsize},
ytick pos=left,
ymin=0.0854166666666667, ymax=0.814583333333333,
yminorgrids,
ytick pos=left,
ytick style={color=black},
every major tick/.append style={major tick length=2pt, black},
ytick={
0.12,
0.195,
0.267,
0.338,
0.413,
0.487,
0.558,
0.63,
0.705,
0.78  
},
yticklabels={ , , , , , , , }
]
\addplot graphics [includegraphics cmd=\pgfimage,xmin=0.09487179485000001, xmax=0.50512820515, ymin=0.0854166666666667, ymax=0.814583333333333] {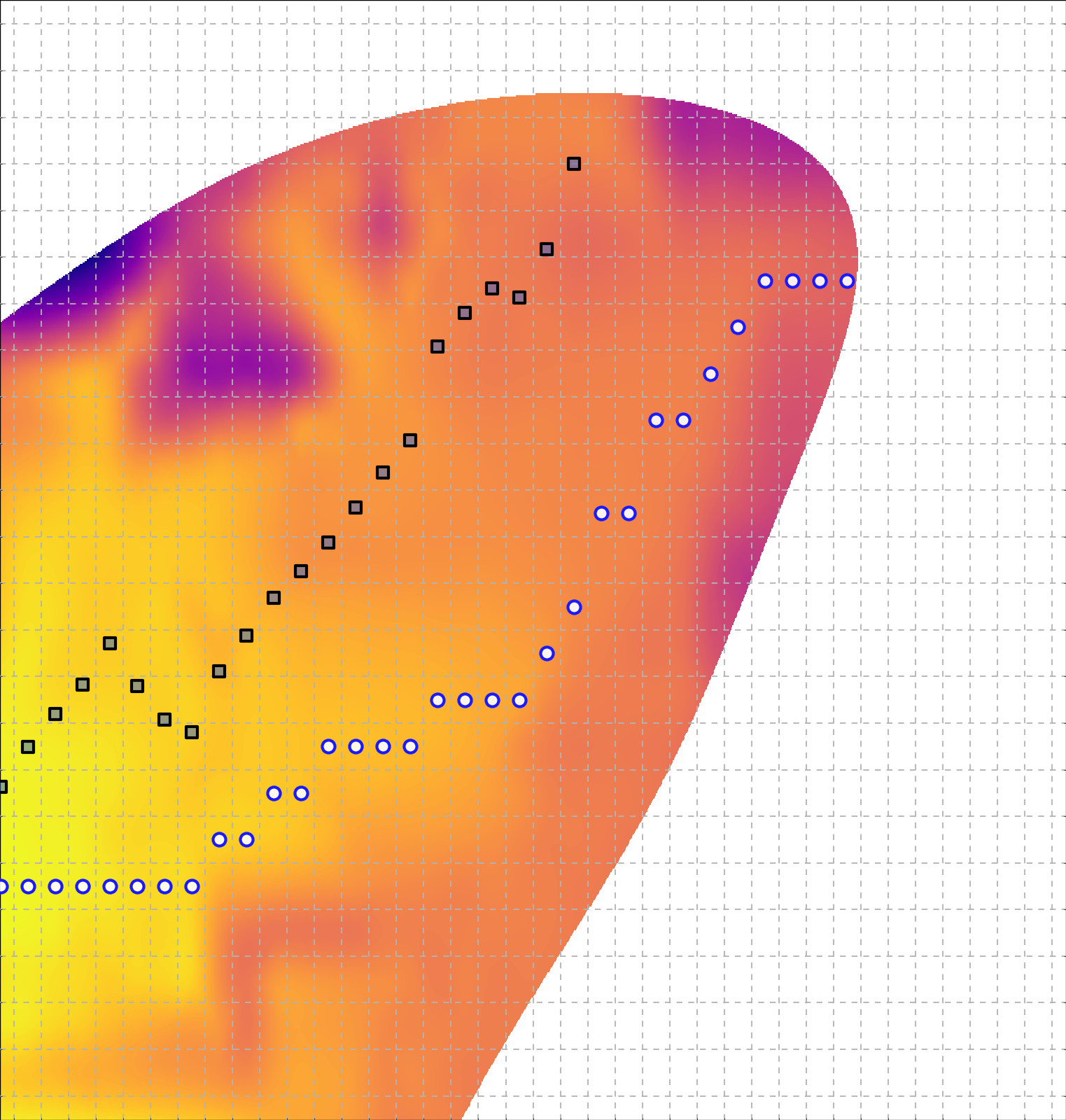};
\addplot graphics [includegraphics cmd=\pgfimage,xmin=0.355, xmax=0.5036, ymin=0.765, ymax=0.81458333] {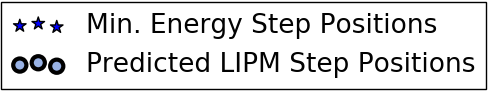};
\end{axis}

\begin{axis}[
at=(plot1.south east),
anchor=south west,
axis on top,
font=\tiny,
width=0.325\textwidth,
height=1.22\textwidth,
xmin=0, xmax=1,
ymin=0, ymax=1,
yminorgrids,
ytick pos=right,
xmajorticks=false,
ytick style={color=black},
ytick={
0.00530573,
0.13351371,
0.26172169,
0.38992968,
0.51813766,
0.64634564,
0.77455362,
0.9027616
},
ylabel style={font=\scriptsize},
ylabel={Measured Torque [\SI{}{\newton\metre}]},
y label style={at={(1.45,0.5)}},
yticklabel style = {rotate=90.0,yshift=0.85mm,xshift=2mm}, 
yticklabels={
10000,
12500,
15000,
17500,
20000,
22500,
25000,
27500 
},
every major tick/.append style={major tick length=2pt, black},
]
\addplot graphics [includegraphics cmd=\pgfimage,xmin=0, xmax=1, ymin=0, ymax=1] {figs/stepping_maps/v4_paper_ready/colorbar_standalone_vertical_no_ticks.png};
\end{axis}
\end{tikzpicture}
}
                \vspace{-5.5mm}
                \caption{}
                \label{fig:lipmtorquecomparison}
            \end{subfigure}%
            \begin{subfigure}{0.305\textwidth}%
                \centering
\raisebox{0.9mm}{
\hspace{8.0mm}
\begin{tikzpicture}
\begin{axis}[
name=plot1,
axis on top,
font=\tiny,
height=1.22\textwidth,
width=1.22\textwidth,
xmin=0.09487179485000001, xmax=0.50512820515,
xminorgrids,
xtick pos=left,
xtick style={color=black},
xtick={
0.1,
0.13636364,
0.17272727,
0.20909091,
0.24545455,
0.28181818,
0.31818182,
0.35454545,
0.39090909,
0.42727273,
0.46363636,
0.5 
},
xticklabel style = {rotate=90.0},
xticklabels={
0.10,
0.14,
0.17,
0.21,
0.25,
0.28,
0.32,
0.35,
0.39,
0.43,
0.46,
0.50
},
ytick pos=left,
xlabel={Initial CoM Vel [m/s]},
x label style={at={(0.5,-0.08)}},
xlabel style={font=\scriptsize},
ymin=0.0854166666666667, ymax=0.814583333333333,
yminorgrids,
ytick pos=left,
ytick style={color=black},
every major tick/.append style={major tick length=2pt, black},
ytick={
0.12,
0.195,
0.267,
0.338,
0.413,
0.487,
0.558,
0.63,
0.705,
0.78  
},
yticklabels={ , , , , , , , }
]
\addplot graphics [includegraphics cmd=\pgfimage,xmin=0.09487179485000001, xmax=0.50512820515, ymin=0.0854166666666667, ymax=0.814583333333333] {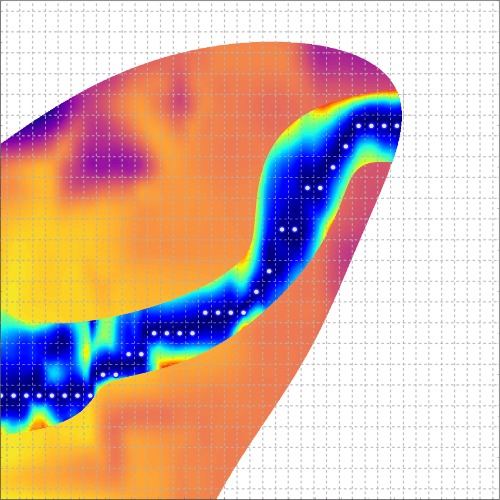};
\addplot graphics [includegraphics cmd=\pgfimage,xmin=0.355, xmax=0.5036, ymin=0.782, ymax=0.81458333] {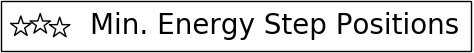};
\end{axis}
\begin{axis}[
at=(plot1.south east),
anchor=south west,
axis on top,
font=\tiny,
width=0.325\textwidth,
height=1.22\textwidth,
xmin=0, xmax=1,
ymin=0, ymax=1,
yminorgrids,
ytick pos=top,
xmajorticks=false,
ytick style={color=black},
ytick={
0.0,
0.10000916,
0.20001831,
0.30002747,
0.40003663,
0.50004578,
0.60005494,
0.7000641,
0.80007325,
0.90008241,
0.99
},
ylabel style={font=\scriptsize},
ylabel={Deviation From Minimum Torque \([\%]\)},
y label style={at={(1,0.5)}},
yticklabel shift=-1mm,
yticklabels={
0,
1,
2,
3,
4,
5,
6,
7,
8,
9,
10
},
every major tick/.append style={major tick length=2pt, black},
]
\addplot graphics [includegraphics cmd=\pgfimage,xmin=0, xmax=1, ymin=0, ymax=1] {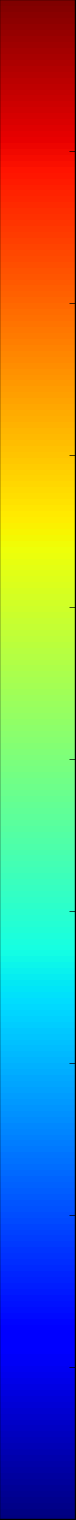};
\end{axis}
\end{tikzpicture}
}
                \vspace{-5.5mm}
                \caption{}
                \label{fig:torquediff}
            \end{subfigure}%
            \caption{
                    Maps of the measured joint torques during the swing phase to reach each position, optimal step locations are marked with stars: (a) a regression model is fitted to the optimal step positions and a region which a maximum \(5\%\) deviation from optimal is shown, (b) energy optimal step positions are compared to predictions from the LIPM model, (c) the heatmap shows which step positions vary from the optimal position by up to \(10\%\) for the same initial CoM velocity.
                }
        \end{figure*}
        
    \subsection{Parameter Optimization - Phase 1}
    \label{sec:parametertuning}
        \subsubsection{Bayesian Optimization}
        \label{sec:bayesianoptimization}
            Phase 1 of our pipeline, shown in the blue box in Figure \ref{fig:systemoverviewdiagram}, optimizes a set trajectory generation parameters, \(\mathbf{p}\), for pairs of initial conditions. Algorithm \ref{alg:bayesoptalg} shows this process: for every initial CoM velocity \(\dot{x}_0\!\in\!\mathbf{\dot{X}}_0\) and every desired step position \(s_{des}\!\in\! \mathbf{S}_{des}\), a set of parameters \(\mathbf{p}\) is optimized via BO, such that for each pair, \([\dot{x}_0, s_{des}]\), is mapped to a set of parameters and stored in the mapping \(\mathscr{G}\).
            
            In total, \(150\) sets of parameters were optimized, \(15\) initial CoM velocities, \(10\) step positions, with 170 BO iterations for each (100 random, 70 Bayesian) using objective Eq.\ref{eq:costfunction}.
            
        \subsubsection{Parameter Interpolation}
        \label{sec:interpolation}
            During optimization, the discrete points which were optimized represented a tiny proportion of the state space, so we interpolate between optimized points to create a continuous mapping \(\mathscr{G}\) (Eq.\ref{eq:mappingfunctiondef}) from arbitrary values of \(\dot{x}_0\) and \(s_{des}\) to a set of interpolated, optimized parameters \(\mathbf{p}^*\). We used element-wise grid interpolation to output a set of linearly interpolated parameter values between the nearest 4 sets of optimized parameters for any given pair of initial conditions.
            
    \subsection{Reachability Maps - Phase 2}
    \label{sec:feasibilitymaps}
        The purpose of phase 2, shown in the orange box in Figure \ref{fig:systemoverviewdiagram}, is to simulate stepping motions from a greater number initial conditions than in phase 1 to produce dense reachability maps. This process is similar to phase 1, but instead of tuning parameters we query the optimal parameters from mapping function \(\mathscr{G}\) and pass these to the trajectory generation functions. Dense maps have three purposes: testing interpolation of optimized parameters, measuring the reachable space of the swing foot, and accurate measurement of joint torque during the swing phase of each step position.
        
        We used \(1000\) initial condition pairs (\(40\) initial CoM velocities, \(20\) desired step positions) to sample the mapping function, compared to \(150\) in phase 1. Initial velocities were between \(\dot{\mathbf{X}}_0 = [\SI{0.1}{\metre\per\second},\dots,\SI{0.5}{\metre\per\second}]\) at \SI{0.017}{\metre\per\second} intervals and desired step positions: \(\mathbf{S}_{des} = [\SI{0.1}{\metre},\dots,\SI{0.8}{\metre}]\) at \SI{0.029}{\metre} intervals. A dynamic simulation episode was executed for each initial condition pair using parameters \(\mathbf{p}^* = \mathscr{G}(\dot{x}_{0_{i}},s_{des_{j}})\forall \: \dot{x}_{0_{i}}\! \in \dot{\mathbf{X}}_0, s_{des_{j}}\! \in \mathbf{S}_{des}\) Eq.\ref{eq:mappingfunctiondef}.
        
        Simulation episodes returned two values: a binary value to denote if the motion was successful, and the integral of joint torques, \(\boldsymbol{\tau}\), using Eq.\ref{eq:torqueobjectivedef}. As a result, each pair of initial conditions has an associated reachability value and measured torque value that we use to build the reachability maps.
            \begin{algorithm}[t]
            \footnotesize
                \SetKwInOut{Input}{input}
                \SetKwInOut{Output}{output}
                \Input{List of Sampled CoM velocities:\(\mathbf{\dot{X}}_0\) \\and desired step positions:\(\mathbf{S}_{des}\)}
                \Output{Mapping function \(\mathscr{G}\) (Eq. \ref{eq:mappingfunctiondef})}
                    \For{each CoM initial velocity \((\dot{x}_0)\) in \(\mathbf{\dot{X}_0}\)}{
                        \For{each step position \((s_{des})\) in \(\mathbf{S}_{des}\)}{
                            \For{each i in BayesOptIterations}{
                                \(\mathbf{p}(i) \leftarrow\) BayesOpt(Eq.\ref{eq:costfunction}) \\
                                objective \(\leftarrow\) DynamicSim(\(\dot{x}_0,s_{des},\mathbf{p}(i)\))
                            }
                        \(\mathscr{G}(\dot{x}_0,s_{des}\)) \(\leftarrow \underset{\mathbf{p}}{\argmax}\)(objective)
                        }
                    }
            \caption{Bayesian Parameter Optimization}
            \label{alg:bayesoptalg}
            \end{algorithm}

        \subsubsection{Binary Reachability Map}
        \label{sec:binaryreachabilitymap}
            A binary reachability map shows which step positions can be reached from initial condition pairs, shown in Figure \ref{fig:svmmap} where green cells denote success and red failure. Parameters returned by the mapping \(\mathscr{G}\) lead to successful stepping over much of the map, but in at the top left map success is more noisy. It is likely that at the extremes of the map, the range of parameters that lead to successful stepping is more narrow than at other points and interpolated parameters do not fall within this range. Optimizing more parameters would give better coverage, but would take longer to tune, we instead trim noisy regions from the map.
                        
            We train an Support Vector Machine (SVM) model to separate the reliable step locations from the noisy areas. Figure \ref{fig:svmmap} shows the support vector overlaid on the reachability map. We then built a high resolution representation of the safe stepping area in Figure \ref{fig:reachvalidationmap} by querying the SVM model with new initial condition pairs. We used a 3rd order SVM with a radial basis function kernel, reachable points had a weight of 1 and unreachable locations had a weight of 14.

        \subsubsection{Measured Torque Maps}
        \label{sec:measuredtorquemaps}
            The torque measured during the swing phase of motions from each initial condition pair is shown in Figure \ref{fig:torquemap}, where each point in the trimmed reachability map is colored according to the integral of joint torques (Eq.\ref{eq:torqueobjectivedef}); darker colors denote higher measured torque. Step positions with the lowest measured torque for each sampled initial velocity are marked with a star. The distribution of measured torque patterns is highly nonlinear, but the pattern of steps with minimum measured torque forms a simple trend which can be used for footstep prediction. We consider the energy-optimal stepping positions to be those with the lowest measured torque integral for all joints.
            
        \begin{figure*}[ht]
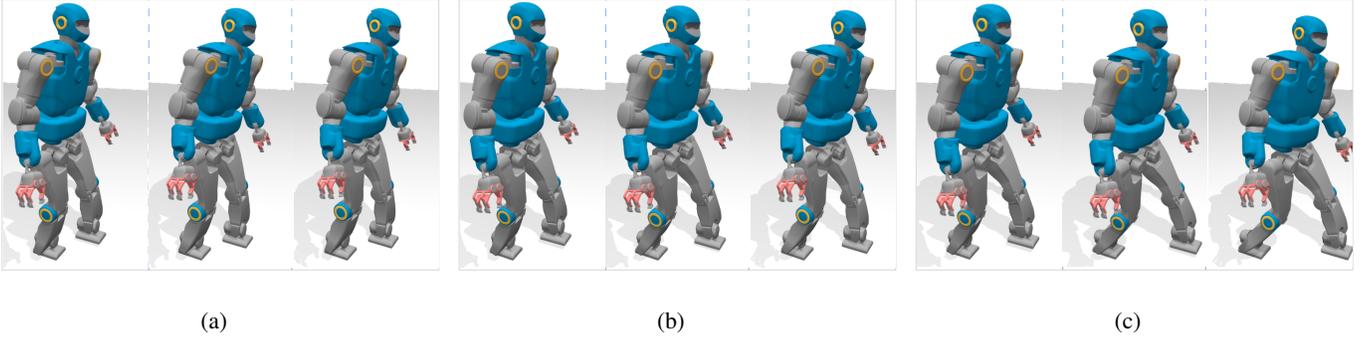

        \vspace{-6mm}
            \begin{subfigure}{0.33\textwidth}
                \timeline%
                        {v4_blender_auto_snapshots/vel_0.164_s_0.2485878536/frame0145_trimmed_480.png}%
                        {v4_blender_auto_snapshots/vel_0.164_s_0.2485878536/frame0190_trimmed_480.png}%
                        {v4_blender_auto_snapshots/vel_0.164_s_0.2485878536/frame0850_trimmed_480.png}
                \caption{}
                \label{sec:validationsnapshots1}
            \end{subfigure}
            \begin{subfigure}{0.33\textwidth}
                \timeline%
                        {v4_blender_auto_snapshots/vel_0.292_s_0.408753569489/frame0115_trimmed_480.png}%
                        {v4_blender_auto_snapshots/vel_0.292_s_0.408753569489/frame0145_trimmed_480.png}%
                        {v4_blender_auto_snapshots/vel_0.292_s_0.408753569489/frame0850_trimmed_480.png}
                \caption{}
                \label{sec:validationsnapshots2}
            \end{subfigure}
            \begin{subfigure}{0.33\textwidth}
                \timeline%
                        {v4_blender_auto_snapshots/vel_0.398666666667_s_0.616582462807/frame0115_trimmed_480.png}%
                        {v4_blender_auto_snapshots/vel_0.398666666667_s_0.616582462807/frame0145_trimmed_480.png}%
                        {v4_blender_auto_snapshots/vel_0.398666666667_s_0.616582462807/frame0850_trimmed_480.png}
                \caption{}
                \label{sec:validationsnapshots3}
            \end{subfigure}
            \caption{
                    Validation of stepping motions using our automatic step selection validation. Initial CoM velocities \SI{0.164}{\metre\per\second}, \SI{0.292}{\metre\per\second},  \SI{0.398}{\metre\per\second}, were mapped to step positions: \SI{0.248}{\metre}, \SI{0.409}{\metre}, \SI{0.617}{\metre} for (a), (b) and (c) respectively.
            }
            \label{fig:validationsnapshotsmain}
        \end{figure*}
    
        \subsubsection{Minimum Energy Step Selection}
        \label{sec:minimumenergystepselection}
            The relationship between initial CoM Velocity and energy-optimal step positions can be modeled using a simple \(4^{\text{th}}\) order polynomial regression, which can quickly approximate energy-optimal step positions given an initial CoM Velocity, as in Eq.\ref{eq:distilledmappingfunction}. Figure \ref{fig:torquemap} shows this model captures the minimal energy step positions trends with a mean error of \SI{216.12}{\newton\metre} (StD.\SI{=287.84}{\newton\metre}, Min.\SI{=0}{\newton\metre}, Max.\SI{=1159.39}{\newton\metre}). This gives us a quick method for selecting energy efficient step positions given the initial CoM velocity of the robot.
            
            Alternatively, energy-optimal positions can be captured more closely using higher order polynomials, which would be slower than our method, or, a set of three linear models could also approximate these step locations less accurately, but more quickly by fitting at the start, end, and connecting them to approximate the middle points.
    
    \subsection{Query of Mapping}
    \label{sec:validation}
        After the mappings have been built, we can query them in order to perform push recovery, this is shown in the green box in Figure \ref{fig:systemoverviewdiagram}, where we combine the mapping functions from the previous two phases. Therefore, for a given CoM velocity, an energy optimal step position is output by the mapping \(s_{des}* = \mathscr{K}(\dot{x}_0)\) (Eq.\ref{eq:distilledmappingfunction}), which is then used to generate the stepping parameters to produce this motion using the mapping \(\mathbf{p}^* = \mathscr{G}(\dot{x}_0,s_{des}^*)\) (Eq. \ref{eq:mappingfunctiondef}). This allows us to automatically generate stepping for a given initial CoM velocity during push motions.
        
        Since the parameters in phase 1 are optimized for set of discrete points, in the next section we run a series of validation tests to show that motions are still produced reliably given continuous initial conditions. In these validation tests, we test the mapping function \(\mathscr{G}\) under a range of initial conditions and the mapping \(\mathscr{K}\) to test the energy optimal step selection. In each case, initial conditions passed to one or both of the mapping functions and their outputs are used to run an episode of the dynamic simulation.
        
\section{Results}
\label{sec:results}
    \subsection{Reachability Map Validation}
    \label{sec:reachabilityvalidation}
        We generated \(1000\) random pairs of initial conditions, \([\dot{x}_0,s_{des}]\), inside the safe stepping region in (Figure \ref{fig:reachvalidationmap}) and ran an episode of dynamic simulation using the mapping \(\mathscr{G}\). For each pair, stepping was dynamic simulated using parameters from the mapping \(\mathscr{G}\) which was judged successful or unsuccessful according to the termination conditions in Section \ref{sec:experimentalsetup}. Figure \ref{fig:reachvalidationmap} shows the initial conditions in the test set projected onto the safe reachability map where successful trials are marked in lighter green and all trials were successful.
    
    \subsection{Energy-Optimal Step Selection Validation}
    \label{sec:stepselectionvalidation}
        Energy optimal step selection was validated with \(150\) initial CoM velocities, randomly generated within the safe region (\(0.1:\SI{0.43}{\metre\per\second}\)). A dynamic simulation episode was executed for each point, using an energy-optimal step position \(s_{des}^*\) (Section \ref{sec:minimumenergystepselection}). All trials were successful; Figure \ref{fig:validationsnapshotsmain} shows snapshots of these motions, with more in the attached video.
        
\begin{figure}[ht]
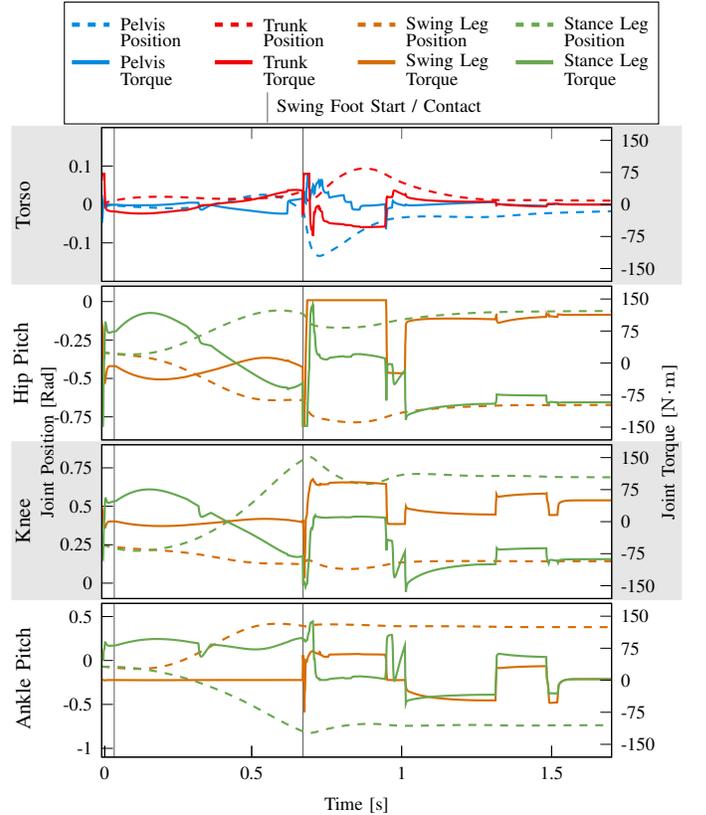

                \include{figs/trajectory_plots/joint_plot_variables}
                \hspace{3mm}
                \begin{subfigure}{\columnwidth}
                    \centering
                        \vspace{-4mm}
                        \begin{tikzpicture}
    \node[] at (0,0.19) {};
    \draw[] (0.0,-0.82) rectangle(7.9,0.8);
    

    \draw[ultra thick,dashed, pelvisposcolor](0.1,0.5) -- (0.6,0.5) node [black,right] {\scriptsize{Pelvis}};
    \draw[ultra thick, black!0](0.1,0.3) -- (0.6,0.3) node [black,right] {\scriptsize{Position}};
    
    \draw[ultra thick, dashed, trunkposcolor](2.0,0.5) -- (2.5  ,0.5) node [black,dashed,right] {\scriptsize{Trunk}};
    \draw[ultra thick, black!0](2.0,0.3) -- (2.5  ,0.3) node [black,right] {\scriptsize{Position}};
    
    \draw[ultra thick, dashed, swingposcolor](3.9,0.5) -- (4.4  ,0.5) node [black,dashed,right] {\scriptsize{Swing Leg}};
    \draw[ultra thick, black!0](3.9,0.3) -- (4.4  ,0.3) node [black,right] {\scriptsize{Position}};
    
    \draw[ultra thick, dashed, stanceposcolor](6.0,0.5) -- (6.5  ,0.5) node [black,dashed,right] {\scriptsize{Stance Leg}};
    \draw[ultra thick, black!0](6.0,0.3) -- (6.5,0.3) node [black,right] {\scriptsize{Position}};

    \draw[ultra thick, pelvisposcolor](0.1,0.0) -- (0.6,0.0) node [black,right] {\scriptsize{Pelvis}};
    \draw[ultra thick, black!0](0.1,-0.25) -- (0.6,-0.25) node [black,right] {\scriptsize{Torque}};
    
    \draw[ultra thick, trunkposcolor](2.0,0.0) -- (2.5  ,0.0) node [black,right] {\scriptsize{Trunk}};
    \draw[ultra thick, black!0](2.0,-0.25) -- (2.5  ,-0.25) node [black,right] {\scriptsize{Torque}};
    
    \draw[ultra thick, swingposcolor](3.9,0.0) -- (4.4  ,0.0) node [black,right] {\scriptsize{Swing Leg}};
    \draw[ultra thick, black!0](3.9,-0.25) -- (4.4  ,-0.25) node [black,right] {\scriptsize{Torque}};
    
    \draw[ultra thick, stanceposcolor](6.0,0.0) -- (6.5  ,0.0) node [black,right] {\scriptsize{Stance Leg}};
    \draw[ultra thick, black!0](6.0,-0.25) -- (6.5,-0.25) node [black,right] {\scriptsize{Torque}};

    \draw[ultra thick, \swingstartcolor,\swingstartlinestyle](2.7,-0.4) -- (2.7,-0.8) node [black,midway,right] {\scriptsize{Swing Foot Start / Contact}};
\end{tikzpicture}
                        \vspace{-2mm}
                \end{subfigure}
                \par
                \begin{subfigure}{0.025\textwidth}%
                    \begin{tikzpicture}
    
    \node[rotate=90] at (0,0) {};
    \filldraw[gray!20] (-0.15,\boundboxstartveritcalA) rectangle(8.75,\boundboxstartveritcalA-\boundboxheight);
    \filldraw[gray!20] (-0.15,\boundboxstartveritcalA-\boundboxheight*2) rectangle(8.75,\boundboxstartveritcalA-\boundboxheight*3);
    
    \node[rotate=90] at (0,\boundboxstartveritcalA-\boundboxheight*0.5) {\footnotesize{Torso}};
    \node[rotate=90] at (0,\boundboxstartveritcalA-\boundboxheight*1.5) {\footnotesize{Hip Pitch}};
    \node[rotate=90] at (0,\boundboxstartveritcalA-\boundboxheight*2.5) {\footnotesize{Knee}};
    \node[rotate=90] at (0,\boundboxstartveritcalA-\boundboxheight*3.5) {\footnotesize{Ankle Pitch}};
    
    \node[rotate=90,align=center] at (0.34,\boundboxstartveritcalA-\boundboxheight*2) {\scriptsize{Joint Position [Rad]}};
    \node[rotate=90,align=center] at (8.6,\boundboxstartveritcalA-\boundboxheight*2) {\scriptsize{Joint Torque [\SI{}{\newton\metre}]}};
\end{tikzpicture}
                \end{subfigure}%
                \begin{subfigure}{0.5\textwidth}%
                    \hspace{1.225mm}
                    \begin{subfigure}{\textwidth}%
                        \input{figs/trajectory_plots/combined_joint_plot_1}
                    \end{subfigure}\hfill%
                    \par\hspace{0mm}
                    \begin{subfigure}{\textwidth}%
                        \input{figs/trajectory_plots/combined_joint_plot_2}
                    \end{subfigure}\hfill%
                    \par\hspace{0.825mm}
                    \begin{subfigure}{\textwidth}%
                        \input{figs/trajectory_plots/combined_joint_plot_3}
                    \end{subfigure}\hfill%
                    \par\hspace{1.225mm}
                    \begin{subfigure}{\textwidth}%
                        \input{figs/trajectory_plots/combined_joint_plot_4}
                    \end{subfigure}\hfill%
                \end{subfigure}%
                \caption{
                            Joint profiles in the torso and legs during one validation stepping motion in Figure \ref{sec:validationsnapshots3}.
                        }
                \label{fig:jointpositions}
                \vspace{-2mm}
            \end{figure}

        Figure \ref{fig:jointpositions} shows the measured joint positions and torques of the torso (trunk and pelvis), hip pitch, knee and ankle pitch joints in both legs during the validation shown in Figure \ref{sec:validationsnapshots3}. During the swing phase, the measured torque in the swing leg joints is lower than in the stance leg since it serves as a rigid pivot for the rest of the body. The landing impact is also clear in the measured torque, after the swing foot makes contact with the ground and since this is close to the actuation limits it can help explain the noisy results in Figure \ref{fig:svmmap}. Comparing actuation profiles quasi-statically stable walking to explore the effects of dynamic stepping on the measured torque at the joints would be an interesting direction for future work.
        
        \begin{figure}[t]
                \begin{subfigure}{\columnwidth}%
                    \hspace{-3.3mm}
\begin{tikzpicture}

\definecolor{color0}{rgb}{0.12156862745098039, 0.4666666666666667, 0.7058823529411765}
\definecolor{color1}{rgb}{1.0, 0.4980392156862745, 0.054901960784313725}
\definecolor{color2}{rgb}{0.17254901960784313, 0.6274509803921569, 0.17254901960784313}
\definecolor{color3}{rgb}{0.8392156862745098, 0.15294117647058825, 0.1568627450980392}
\definecolor{color4}{rgb}{0.5803921568627451, 0.403921568627451, 0.7411764705882353}
\definecolor{color9}{rgb}{0.5490196078431373, 0.33725490196078434, 0.29411764705882354}
\definecolor{color6}{rgb}{0.8901960784313725, 0.4666666666666667, 0.7607843137254902}
\definecolor{color7}{rgb}{0.4980392156862745, 0.4980392156862745, 0.4980392156862745}
\definecolor{color8}{rgb}{0.7372549019607844, 0.7411764705882353, 0.13333333333333333}
\definecolor{color5}{rgb}{0.09019607843137255, 0.7450980392156863, 0.8117647058823529}

\begin{axis}[
axis on top,
legend cell align={left},
tick pos=both,
xtick pos=bottom,
ytick pos=left,
legend columns=5,
legend style={at={(0.015,1)}, anchor=south west},
xlabel={Initial CoM Velocity [m/s]},
xmin=0.092, xmax=0.508,
xtick style={color=black},
xtick={
0.1,
0.13636364,
0.17272727,
0.20909091,
0.24545455,
0.28181818,
0.31818182,
0.35454545,
0.39090909,
0.42727273,
0.46363636,
0.5 
},
xticklabels={
0.1,
0.14,
0.17,
0.21,
0.25,
0.28,
0.32,
0.35,
0.39,
0.43,
0.46,
0.5 
},
ylabel={Time [s]},
ytick={10,20,30,40},yticklabels={600,1200,1800,2400},
y label style={at={(-0.0579,0.5)}},
ymin=10, ymax=45,
ytick style={color=black},
yticklabel style={/pgf/number format/fixed,/pgf/number format/precision=0,/pgf/number format/fixed zerofill},
                                            scaled y ticks = false,
                                            height=0.5\columnwidth,
                                            width=1.107\columnwidth,
                                            font=\scriptsize,
]
\addplot [thick, color0, mark=*, mark size=2, mark options={solid,draw=black}]
table {%
0.1 43.5778157313333
0.136363636363636 41.5677174011667
0.172727272727273 40.7915647665
0.209090909090909 39.5565615853333
0.245454545454545 36.8408182303333
0.281818181818182 26.6383135358333
0.318181818181818 24.7336436151667
0.354545454545455 24.618403236
0.390909090909091 26.3749352495
0.427272727272727 24.573369968
0.463636363636364 24.4793253341667
0.5 22.1873216828333
};
\addlegendentry{0.10[m]}
\addplot [thick, color1, mark=*, mark size=2, mark options={solid,draw=black}]
table {%
0.1 42.2326702158333
0.136363636363636 40.8403943498333
0.172727272727273 38.9686572193333
0.209090909090909 37.2318098306667
0.245454545454545 36.247350665
0.281818181818182 32.7192091186667
0.318181818181818 32.5994765678333
0.354545454545455 22.7473317186667
0.390909090909091 24.3335573315
0.427272727272727 26.0585054
0.463636363636364 25.2922190506667
0.5 27.5691516001667
};
\addlegendentry{0.18[m]}
\addplot [thick, color2, mark=*, mark size=2, mark options={solid,draw=black}]
table {%
0.1 38.6720529
0.136363636363636 38.446191748
0.172727272727273 37.5166338681667
0.209090909090909 32.6066390315
0.245454545454545 38.2640163501667
0.281818181818182 33.8710380315
0.318181818181818 33.4300989826667
0.354545454545455 27.80523355
0.390909090909091 28.7138316631667
0.427272727272727 28.4710229356667
0.463636363636364 26.8980473836667
0.5 26.2412044008333
};
\addlegendentry{0.26[m]}
\addplot [thick, color3, mark=*, mark size=2, mark options={solid,draw=black}]
table {%
0.1 36.6738599498333
0.136363636363636 39.5316706656667
0.172727272727273 36.4205730636667
0.209090909090909 37.5675533493333
0.245454545454545 36.744121317
0.281818181818182 33.3895189841667
0.318181818181818 34.6736528635
0.354545454545455 26.6544636528333
0.390909090909091 28.2045616508333
0.427272727272727 25.0065120021667
0.463636363636364 22.2290964485
0.5 23.370917066
};
\addlegendentry{0.33[m]}
\addplot [thick, color4, mark=*, mark size=2, mark options={solid,draw=black}]
table {%
0.1 32.4379279495
0.136363636363636 36.665405651
0.172727272727273 35.5910236676667
0.209090909090909 34.771300749
0.245454545454545 33.3205893675
0.245454545454545 22.2623257995
0.281818181818182 29.888807269
0.318181818181818 29.1295941193333
0.354545454545455 29.5144319335
0.390909090909091 20.2961938858333
0.427272727272727 20.3052398681667
0.463636363636364 21.9473843693333
0.5 20.8114801805
};
\addlegendentry{0.41[m]}
\addplot [thick, color5, mark=*, mark size=2, mark options={solid,draw=black}]
table {%
0.1 34.9025341153333
0.136363636363636 36.1870869
0.172727272727273 34.925343132
0.209090909090909 30.9261926491667
0.245454545454545 30.1450743318333
0.281818181818182 29.707180798
0.318181818181818 29.5894537686667
0.354545454545455 29.6646348833333
0.390909090909091 28.9800774851667
0.427272727272727 24.7704482675
0.463636363636364 22.3508520841667
0.5 24.8347914178333
};
\addlegendentry{0.49[m]}
\addplot [thick, color6, mark=*, mark size=2, mark options={solid,draw=black}]
table {%
0.1 32.960547483
0.136363636363636 31.7752610643333
0.172727272727273 31.5974908153333
0.209090909090909 29.63347677
0.245454545454545 29.3269244671667
0.281818181818182 29.2189343651667
0.318181818181818 30.4298423171667
0.354545454545455 26.3941507656667
0.390909090909091 26.848443067
0.427272727272727 25.9555516163333
0.463636363636364 24.6797354658333
0.5 23.0856873353333
};
\addlegendentry{0.57[m]}
\addplot [thick, color7, mark=*, mark size=2, mark options={solid,draw=black}]
table {%
0.1 23.5228779833333
0.136363636363636 23.2520407995
0.172727272727273 24.2827726641667
0.209090909090909 26.361746502
0.245454545454545 25.796086963
0.281818181818182 25.9167983333333
0.318181818181818 26.177513349
0.354545454545455 28.0471318483333
0.390909090909091 26.1780764341667
0.427272727272727 26.2386086663333
0.463636363636364 26.2740237831667
0.5 20.4658399145
};
\addlegendentry{0.64[m]}
\addplot [thick, color8, mark=*, mark size=2, mark options={solid,draw=black}]
table {%
0.1 17.2862967015
0.136363636363636 16.5109988331833
0.172727272727273 17.7656610648333
0.209090909090909 20.0603034536667
0.245454545454545 18.9589832861667
0.281818181818182 23.1131520828333
0.318181818181818 24.380361398
0.354545454545455 23.4623681863333
0.390909090909091 25.8090478658333
0.427272727272727 22.2054599681667
0.463636363636364 23.4607957005
0.5 20.4587582151667
};
\addlegendentry{0.72[m]}
\addplot [thick, color9, mark=*, mark size=2, mark options={solid,draw=black}]
table {%
0.1 13.2797606825833
0.136363636363636 12.65709421635
0.172727272727273 13.94298051595
0.209090909090909 16.2464548349333
0.245454545454545 15.6728702664333
0.281818181818182 19.6032490135
0.318181818181818 21.1056794485
0.354545454545455 21.1105957985
0.390909090909091 16.6394775470167
0.427272727272727 19.789060549
0.463636363636364 20.547257483
0.5 22.5319416681667
};
\addlegendentry{0.80[m]}
\end{axis}
\end{tikzpicture}
                    \vspace{-7.5mm}
                    \caption{}
                    \label{fig:timeplot}
                \end{subfigure}%
                \par
                \hspace{4mm}
                \begin{subfigure}{\columnwidth}%
                    \centering
                    \vspace{1.5mm}
                    \input{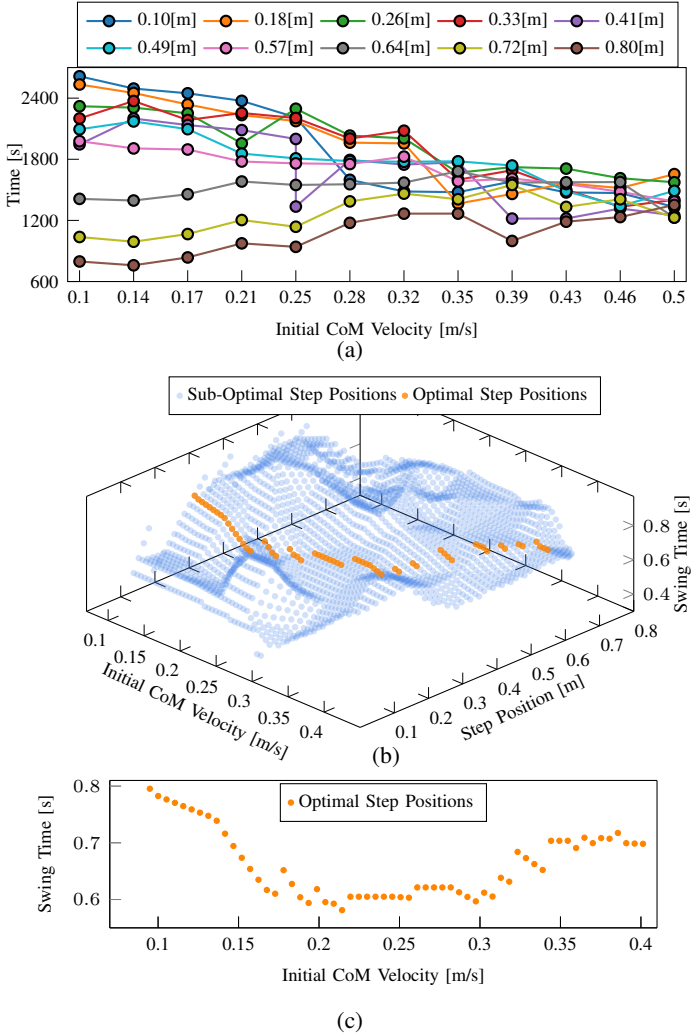}
                    \hspace{-4mm}
                    \vspace{-6mm}
                    \caption{}
                    \label{fig:3dswingplot}
                \end{subfigure}%
                \par
                \begin{subfigure}{\columnwidth}%
                    \centering
                    \begin{tikzpicture}

\definecolor{color0}{rgb}{1,0.52156862745098,0}
\definecolor{color1}{rgb}{0.290196078431373,0.525490196078431,0.909803921568627}

\begin{axis}[
view={0}{0},
legend cell align={left},
tick pos=both,
xlabel={Initial CoM Velocity [m/s]},
ylabel={Step Position [m]},
zlabel={Swing Time [s]},
xmin=0.07, xmax=0.41,
ymin=0, ymax=0.8,
zmin=0.55, zmax=0.81,
ztick pos=left,
xtick pos = bottom,
y label style={rotate=25,anchor=center},
legend style={at={(0.5,0.85)},anchor=center},
xtick style={color=black},
legend columns=2,
xtick={0.1,0.15,0.2,0.25,0.3,0.35,0.4},xticklabels={0.1,0.15,0.2,0.25,0.3,0.35,0.4},
ytick={0.1,0.2,0.3,0.4,0.5,0.6,0.7,0.8},yticklabels={0.1,0.2,0.3,0.4,0.5,0.6,0.7,0.8},
ytick style={color=black},
yticklabel style={/pgf/number format/fixed,/pgf/number format/precision=0,/pgf/number format/fixed zerofill},
                                        scaled y ticks = false,
                                        height=0.4\columnwidth,
                                        width=\columnwidth,
                                        font=\scriptsize
]
\addplot3[only marks,  draw=color0, fill=color0, mark size=1, opacity=1, colormap={mymap}{[1pt]
  rgb(0pt)=(0,0,0.5);
  rgb(22pt)=(0,0,1);
  rgb(25pt)=(0,0,1);
  rgb(68pt)=(0,0.86,1);
  rgb(70pt)=(0,0.9,0.967741935483871);
  rgb(75pt)=(0.0806451612903226,1,0.887096774193548);
  rgb(128pt)=(0.935483870967742,1,0.0322580645161291);
  rgb(130pt)=(0.967741935483871,0.962962962962963,0);
  rgb(132pt)=(1,0.925925925925926,0);
  rgb(178pt)=(1,0.0740740740740741,0);
  rgb(182pt)=(0.909090909090909,0,0);
  rgb(200pt)=(0.5,0,0)
}]
table{%
x                      y                 f(x)
0.09487179485000001 0.2639880952295918 0.7954303992038328
0.10006491396772153 0.2639880952295918 0.7825802937122509
0.10525803308544304 0.2639880952295918 0.7765041880390732
0.11045115220316457 0.2639880952295918 0.7705323400948918
0.11564427132088609 0.2639880952295918 0.7646620937042053
0.12083739043860761 0.2639880952295918 0.7588908821602703
0.1260305095563291 0.2639880952295918 0.7532162244895534
0.13122362867405063 0.2639880952295918 0.747635721901796
0.13641674779177215 0.2639880952295918 0.7389525046751182
0.14160986690949368 0.2639880952295918 0.7160997874247913
0.1468029860272152 0.2639880952295918 0.6943555210632252
0.15199610514493672 0.2639880952295918 0.6736274241650699
0.15718922426265824 0.2639880952295918 0.6538331822718733
0.16238234338037977 0.2639880952295918 0.634899137652692
0.1675754624981013 0.2639880952295918 0.6167591804597317
0.17276858161582279 0.2639880952295918 0.6102494536126484
0.1779617007335443 0.29374999999285717 0.6516582996251181
0.18315481985126583 0.29374999999285717 0.627479371447189
0.18834793896898733 0.29374999999285717 0.603923969055651
0.19354105808670885 0.29374999999285717 0.5938880468958415
0.19873417720443037 0.32351190475612246 0.6182857034974448
0.2039272963221519 0.32351190475612246 0.5953913527415499
0.20912041543987342 0.32351190475612246 0.5926657590962208
0.21431353455759494 0.32351190475612246 0.5812353535110152
0.21950665367531647 0.35327380951938775 0.6049688388719562
0.224699772793038 0.35327380951938775 0.6049688388719562
0.2298928919107595 0.35327380951938775 0.6049688388719561
0.23508601102848103 0.35327380951938775 0.6049688388719562
0.24027913014620256 0.35327380951938775 0.6049688388719562
0.24547224926392402 0.35327380951938775 0.6049658670220384
0.25066536838164555 0.35327380951938775 0.6040954126731553
0.25585848749936707 0.35327380951938775 0.6032275191621499
0.2610516066170886 0.38303571428265304 0.6213199887636013
0.2662447257348101 0.38303571428265304 0.6213208922265373
0.27143784485253164 0.38303571428265304 0.6213217956922822
0.27663096397025316 0.38303571428265304 0.6213226991608362
0.2818240830879747 0.38303571428265304 0.6213136270077618
0.2870172022056962 0.38303571428265304 0.6127438666306202
0.29221032132341773 0.38303571428265304 0.6045743225007382
0.29740344044113926 0.38303571428265304 0.5967835847196795
0.3025965595588608 0.4127976190459184 0.612067592629064
0.3077896786765823 0.4127976190459184 0.6052850206674939
0.31298279779430377 0.4425595238091837 0.6382782141882615
0.3181759169120253 0.4425595238091837 0.6314519892695112
0.3233690360297468 0.5020833333357143 0.6840066293308026
0.32856215514746834 0.5020833333357143 0.6730723537589969
0.33375527426518986 0.5020833333357143 0.6624695284772479
0.3389483933829114 0.5020833333357143 0.6521824303164429
0.3441415125006329 0.5616071428622449 0.7036121742715967
0.34933463161835443 0.5616071428622449 0.7036121695387905
0.35452775073607595 0.5616071428622449 0.703612164805984
0.3597208698537975 0.5616071428622449 0.6908784889476242
0.364913988971519 0.5913690476255102 0.7090770936522112
0.3701071080892405 0.5913690476255102 0.6994208419136522
0.37530022720696204 0.6211309523887755 0.708390101298933
0.38049334632468357 0.6211309523887755 0.7072584573391736
0.3856864654424051 0.6508928571520408 0.7173981159797788
0.3908795845601266 0.6508928571520408 0.699366928990257
0.3960727036778481 0.6508928571520408 0.6986852853510809
0.4012658227955696 0.6508928571520408 0.6982580960679323
};
\addlegendentry{Optimal Step Positions}
\end{axis}

\end{tikzpicture}
                    \caption{}
                    \label{fig:2dswingplot}
                \end{subfigure}%
                \caption{
                        Analysis of pipeline performance and parameter results: (a) Computational time for optimizing parameters for each pair of initial conditions, step positions are denoted by colored traces; (b) 3D plot of swing time for optimal and sub optimal points; (c) 2D projection of swing time for optimal points.
                    }
        \end{figure}
    \subsection{LIPM Comparison}
    \label{sec:lipmcomparison}
        The mapping \(\mathscr{G}\) was then queried forward simulate the LIPM model to predict one step capturable step positions with the same initial conditions, as shown in Figure \ref{fig:lipmtorquecomparison}. After \SI{0.32}{\metre\per\second}, the LIPM prediction is beyond the reachable area for the robot, so are not included. We calculated the error in measured torque between the optimal step positions from our optimization and those from the LIPM prediction, with a mean root squared error of \SI{1364.5}{\newton\metre} (StD.\(=\)\SI{829.6}{\newton\metre}, Min.\(=\)\SI{164.3}{\newton\metre}, Max.\(=\)\SI{2531.9}{\newton\metre}). Comparing our results with the LIPM shows the extent of the difference between the reduced and full dynamic models: the energy cost is higher when using simple models and many of step locations are outside the reachable area.

\subsection{Diversity in Balancing Strategies}
\label{sec:simplicityanddiversity}
        In addition to the simple trend of energy-optimal step positions, a diverse range of near-optimal step locations are highlighted on the map. The highlighted area in Figure \ref{fig:torquemap} shows step positions where energy costs are a maximum of \(5\%\) above optimal for the same initial condition and Figure \ref{fig:torquediff} shows a band of step positions which deviate from measured optimal by a maximum of \(10\%\).
        
        Near-optimal regions form simple heuristics: for low initial CoM velocities, stepping between \SI{22}{\centi\metre} - \SI{30}{\centi\metre} will result in minimal impact on energy efficiency, even if the optimal position is not reached, allowing coarse, yet rapid step selection with trivial changes in energy efficiency. Additionally, since the regions span a range of initial CoM velocities, inaccurate CoM state estimation can still lead to efficient stepping.

    \subsection{Underlying Energy Optimality}
    \label{sec:underlyingoptimality}
        Near-optimal efficiency regions suggest an underlying optimality in dynamic stepping gives us insight into human step selection. If similar regions exist in dynamic multi-body systems in general, humans can learn similar heuristics and use them for simple, rapid, near-optimal step selection. Given sensing delays in humans and inaccuracies in sensing disturbances, having near-optimal heuristics that work despite delays would be beneficial to developing humans.
        
        Figure \ref{fig:humansteppos}, shows a clustering of selected steps similar to that in highlighted area in Figure \ref{fig:torquemap}, but with a higher range of initial CoM velocities. Humans are able to withstand higher magnitudes using foot tilting behavior. This is a limitation of our robot control, which does not consider underactuated foot tilting control, hindering the range of feasible step positions. 
        
        \balance
        We also gain insight into energy optimality by looking at the optimized step parameters. Total swing time can be calculated from the optimal parameters (\(t_{swing\_start}+s_{des}/s_{speed}\)), and plotted in Figure \ref{fig:3dswingplot}), showing all initial condition pairs and their optimized swing time, with only optimal points projected into 2D in Figure \ref{fig:2dswingplot}. This shows a piece-wise relationship between initial CoM velocity and optimal swing time, where swing time initially drops, levels off, then rises as the initial CoM velocity increases. The majority of the swing time effect is caused by the swing speed parameter, \(s_{swing\_speed}\), with only minor changes induced by the swing start time, \(t_{swing\_start}\), as shown by the optimal ranges in Figure \ref{tab:optvariables}.
        
    \subsection{Analysis of Computation Time}
    \label{sec:timinganalysis}
        Training used an Intel Core i7-8700k with 12 cores (6 physical), 32GB RAM, Ubuntu 16.04 and Pinocchio 2.5.0. The pipeline is parallelized, with each core optimizing all step positions for one initial CoM velocity, and took around 5 hours, reachability map building takes around 40 minutes. Parameters for one swing foot were used for the opposite foot.
        
        Figure \ref{fig:timeplot} shows the computation time for parameter optimization for each pair of initial conditions, with \(170\) episodes for each pair. For larger initial CoM velocities and larger step distances, the number of early terminations leads to quicker computation. Parallelization scales linearly with the number of cores, where large-scale distribution, with one core per initial condition pair, would lead to \SI{45}{\minute} optimization.
        
        Swing foot trajectories are generated in \SI{0.5}{\milli\second}, and CoM trajectories in \SI{1.4}{\second}, due to the nonlinear optimization formation used for generation, but can be reduced to \SI{0.5}{\milli\second} using the same generation as the swing foot. Querying the energy-optimal step selection takes \SI{0.13}{\milli\second}.

\section{Discussion}
\label{sec:discussion}
    
    The study of energy efficient locomotion is a complex process which involves the whole-body dynamics and control together as a whole and requires complex optimization, as well as global search of global energy-optimal step location and timing, which are all highly nonlinear. The complex interplay between multi-body dynamics, control and gait parameters is shown by the optimal gait parameters in Figure \ref{fig:3dswingplot}, where this nonlinear relationship and non-smooth gradient would cause standard gradient search methods to get stuck in the local minima. However, the mapping between initial conditions and key gait parameters, such as step location and swing time, suggests that the gait parameters can be represented by piece-wise approximations, as shown in Figures \ref{fig:torquemap} \& \ref{fig:2dswingplot}. This indicates that despite the complexity of the whole process, this nonlinear relationship in human gait can be possibly learned by humans by prior trials and experience.
    
    Additionally, balance can be recovered using similar step locations and swing times for different initial CoM velocities by trading off energy optimality, which can potentially explain the large variations in step location in human study \cite{mcgreavy2020unified}.
    
\section{Conclusion}
\label{sec:conclusion}
    In this paper, we investigated energy efficient step selection using nonlinear optimization to build offline reachability maps. We identified that selecting energy efficient steps during push recovery or finding a set of diverse stepping regions are difficult to characterize with simple models, and are hard to compute online (see Figure \ref{fig:timeplot}). Hence reachability maps can be used for rapid step selection. Results also give us insight into the possibility and feasibility of diverse step selection for humanoids. In our future work, we plan to extend this pipeline to study energy efficient locomotion in different modes, and implement the Query of Mapping for warm-start solution for online optimization. 
    
\bibliographystyle{IEEEtran}
\bibliography{bibliography}

\vspace{-0.5cm}
\begin{IEEEbiography}[{\includegraphics[width=1in,height=1.25in,clip,keepaspectratio]{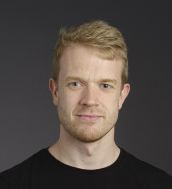}}]{Christopher McGreavy} received his M.Sc in Computational Neuroscience and Cognitive Robotics from the University of Birmingham, and M.Sc.(R) from the University of Edinburgh in Robotics and Autonomous Systems. He is currently a PhD student at the University of Edinburgh with research interests in robotic and human locomotion, enhancing understanding and application of efficient, dynamic walking.
\end{IEEEbiography}

\vspace{-0.5cm}
\begin{IEEEbiography}[{\includegraphics[width=1in,height=1.25in,clip,keepaspectratio]{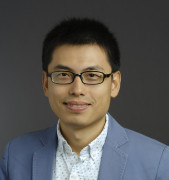}}]{Zhibin (Alex) Li} is an assistant professor at the School of Informatics, University of Edinburgh. He obtained a joint PhD degree in Robotics at the Italian Institute of Technology (IIT) and University of Genova in 2012. His research interests are in creating intelligent behaviors of dynamical systems with human comparable abilities to move and manipulate by inventing new control, optimization and deep learning technologies.
\end{IEEEbiography}

\end{document}